\documentclass[11pt]{article}

\usepackage[final]{acl}

\usepackage{times}
\usepackage{latexsym}

\usepackage[T1]{fontenc}

\usepackage[utf8]{inputenc}

\usepackage{microtype}

\usepackage{inconsolata}

\usepackage{graphicx}

\usepackage{xcolor} 
\usepackage{subcaption}
\usepackage{todonotes}

\usepackage{xspace}
\newcommand{\model}{\textsc{COREA}\xspace} 
\usepackage{amsmath, amsfonts, amssymb}
\usepackage{booktabs} 
\usepackage{multirow} 
\usepackage{makecell} 

\usepackage{xcolor}
\usepackage{colortbl}

\usepackage[utf8]{inputenc}
\usepackage{tcolorbox}

\definecolor{darkgray}{RGB}{40,40,40}
\definecolor{lightgray}{RGB}{230,230,230}
\definecolor{purple}{RGB}{150,100,130}
\definecolor{lightpurple}{RGB}{240,220,240}
\definecolor{darkblue}{RGB}{20,40,100}
\definecolor{lightblue}{RGB}{200,220,255}
\definecolor{darkgreen}{RGB}{0,100,0}
\definecolor{lightgreen}{RGB}{200,255,200}

\title{
Confidence-Calibrated Small-Large Language Model Collaboration for Cost-Efficient Reasoning
}

\author{
 \textbf{Chuang Zhang\textsuperscript{1}},
 \textbf{Zizhen Zhu\textsuperscript{2}},
 \textbf{Yihao Wei\textsuperscript{1}},
 \textbf{Bing Tian\textsuperscript{1}},
\\
 \textbf{Junyi Liu\textsuperscript{1}},
 \textbf{Henan Wang\textsuperscript{1}},
 \textbf{Xavier Wang\textsuperscript{1}},
 \textbf{Yaxiao Liu\textsuperscript{1}}
\\
 \textsuperscript{1}Amazon Web Services \quad
 \textsuperscript{2}Tsinghua University
\\
\{zhangchg, weiyihao, tiabingn, liujunyi, henanwan, wangxy, liuyaxia\}@amazon.com
\\
zzz23@mails.tsinghua.edu.cn
}

\begin{document}
\maketitle

\begin{abstract}

Large language models (LLMs) demonstrate superior reasoning capabilities compared to small language models (SLMs), but incur substantially higher costs. We propose \textbf{CO}llaborative \textbf{REA}soner (\model), a system that cascades an SLM with an LLM to achieve a balance between accuracy and cost in complex reasoning tasks. \model first attempts to answer questions using the SLM, which outputs both an answer and a verbalized confidence score. Questions with confidence below a predefined threshold are deferred to the LLM for more accurate resolution. We introduce a reinforcement learning-based training algorithm that aligns the SLM's confidence through an additional confidence calibration reward. Extensive experiments demonstrate that our method jointly improves the SLM's reasoning ability and confidence calibration across diverse datasets and model backbones. Compared to using the LLM alone, \model reduces cost by 21.5\% and 16.8\% on out-of-domain math and non-math datasets,  respectively, with only an absolute pass@1 drop within 2\%. 

\end{abstract}
\section{Introduction}
\label{sec:introduction}

Reasoning ability has emerged as a central performance metric for large language models (LLMs), enabling them to solve complex problems in mathematics, science, and coding, etc. A key strategy for unlocking this capability is the generation of explicit chain-of-thought (CoT) reasoning steps \cite{wei2023chainofthoughtpromptingelicitsreasoning}, often combined with reinforcement learning (RL) or supervised fine-tuning (SFT). While highly effective, this approach introduces significantly higher inference cost due to the verbose reasoning traces, posing substantial barriers to real-world integration at scale \cite{wang2025harnessingreasoningeconomysurvey}.

To address the high inference costs, recent research has explored several approaches. One direction involves training small language models (SLMs) that are more cost-efficient through knowledge distillation \cite{yu2024distillingsystem2system1,chen2024mixeddistillationhelpssmaller}. However, SLMs typically exhibit weaker reasoning capabilities than LLMs \cite{li2025smallmodelsstrugglelearn}. Another line of work focuses on optimizing reasoning efficiency within a single model through techniques such as adaptive computation \cite{snell2024scalingllmtesttimecompute}, latent reasoning compression \cite{hao2024training}, or dynamic reasoning paradigms \cite{sun2024fastbestndecodingspeculative}. While these methods can reduce computational overhead, they require modifying the model architecture or training procedure, and their effectiveness is still limited by the underlying model's capacity. Routing-based approaches have also been proposed to direct queries to appropriately-sized models \cite{ong2024routellmlearningroutellms,aytes2025sketchofthoughtefficientllmreasoning}, yet they typically rely on external classifiers or heuristics that may not accurately capture the model's actual reasoning confidence. These limitations reveal a persistent trade-off: using SLMs alone sacrifices accuracy and robustness on challenging problems, while relying exclusively on LLMs incurs prohibitively high costs.

This contrast motivates a practical question: \textbf{can we combine the cost-efficiency of SLMs with the accuracy and robustness of LLMs?}
Ideally, an SLM should take the first pass—solving problems it can handle confidently—and defer harder problems to a more powerful LLM.
This collaborative approach promises to deliver high overall accuracy while keeping operational costs significantly lower than exclusive LLM usage.
However, implementing such a collaborative system is non-trivial.
The key technical challenge lies in enabling the SLM to recognize its own limitations and make informed decisions about whether to answer or defer. 
Specifically, SLMs currently \textit{do not know what they know, nor do they know what they don't know}~\cite{DBLP:conf/acl/YinSGWQH23,DBLP:conf/emnlp/PratoHPSC24,DBLP:conf/nips/KapoorGRCPBWDGW24}.

In this work, we propose~\textbf{\model} (\textbf{CO}llaborative \textbf{REA}soner), which empowers a self-aware SLM to dynamically coordinate with an LLM during inference. Self-awareness is achieved by calibrating the SLM's verbalized confidence using RL training with a customized reward function. In this system, when the SLM is confident and capable of providing a correct answer, it directly responds to the query; when uncertain, i.e., its self-assessed confidence falls below a threshold, it hands the task over to the LLM for further processing.

To summarize, our main contributions are:

1. We propose \model, an SLM-LLM collaborative system that balances accuracy and cost by enabling the SLM to defer challenging problems to the LLM when confidence is low.

2. We develop an RL training method that combines a verifiable reward with a confidence calibration reward, and demonstrate empirically that it simultaneously improves reasoning ability and confidence calibration.

3. We conduct extensive experiments validating both the effectiveness of our RL training method and the performance of the collaborative system.

\section{Related Work}
\label{sec:related_work}

\subsection{Efficient Reasoning}

LLMs such as GPT-5-thinking~\cite{openai2025gpt5}, Claude 4~\cite{anthropic2025claude4}, and DeepSeek-R1~\cite{deepseekai2025deepseekr1incentivizingreasoningcapability} achieve strong reasoning performance by producing explicit reasoning traces~\cite{li202512surveyreasoning}, but verbose outputs incur high token costs and latency~\cite{wang2025harnessingreasoningeconomysurvey}. Research on efficient reasoning follows three main strategies. \textbf{Single-model optimization} reduces overhead through adaptive computation~\cite{snell2024scalingllmtesttimecompute}, latent reasoning compression~\cite{hao2024training}, or dynamic reasoning~\cite{sun2024fastbestndecodingspeculative}, but requires architectural modifications and remains limited by the model's capacity. \textbf{Enhancing SLMs} via distillation~\cite{yu2024distillingsystem2system1,chen2024mixeddistillationhelpssmaller,deepseekai2025deepseekr1incentivizingreasoningcapability}, SFT~\cite{muennighoff2025s1simpletesttimescaling}, or RL~\cite{li2025limrrlscaling} enables cost-efficient deployment, yet distilled SLMs often struggle on complex problems~\cite{li2025smallmodelsstrugglelearn} and exhibit overconfidence. \textbf{SLM–LLM collaboration} routes queries between models: query-level routing~\cite{ong2024routellmlearningroutellms,aytes2025sketchofthoughtefficientllmreasoning} uses external classifiers that may misestimate difficulty, while fine-grained methods~\cite{akhauri2025splitreasonlearningoffloadreasoning,kim2025guidingreasoningsmalllanguage,DBLP:conf/coling/LeeYTHBC24,lee2025collaborativellminferenceplanning} and advanced cascades~\cite{DBLP:conf/iclr/GuptaNJRMK24,DBLP:journals/corr/abs-2310-03094} defer challenging steps or employ uncertainty-based voting. However, these collaborative approaches typically require additional sampling overhead or external modules, in contrast, we directly calibrate the SLM's internal confidence for reliable routing decisions.

\subsection{Uncertainty and Confidence Calibration}

Accurate self-assessment is critical for SLM–LLM collaboration, yet SLMs often lack reliable confidence calibration. Early studies analyzed prompting strategies and benchmarks for expressing doubt~\cite{DBLP:conf/acl/YinSGWQH23,DBLP:conf/emnlp/PratoHPSC24,DBLP:conf/nips/KapoorGRCPBWDGW24}, revealing that RLHF often produces overconfident models~\cite{DBLP:conf/iclr/LengHZ025}. Research on improving calibration follows three main approaches. \textbf{Eliciting verbalized confidence} through prompt engineering~\cite{yang2025on,DBLP:journals/tmlr/LinHE22} enables models to express uncertainty explicitly, however, calibration quality varies significantly across tasks and templates. \textbf{Introspection-based methods} leverage hidden states or auxiliary classifiers to detect when the model may be incorrect~\cite{DBLP:conf/emnlp/AzariaM23}, still, the performance is generally hard to extend to out-of-distribution tasks. \textbf{Training for calibration} incorporates confidence calibration into the learning objective: like including a Brier score \cite{DBLP:journals/corr/abs-2507-16806, DBLP:conf/emnlp/XuWDL0CG24}, Logarithmic score \cite{stangel2025rewardingdoubtreinforcementlearning} in RL training. We follow the similar training approach, however, we propose several different calibration rewards and make an extensive comparison on these reward designs. 
\section{Methodology}
\label{sec:methodology}

\begin{figure*}[htbp]
\centering
\begin{subfigure}[t]{0.52\textwidth}
\centering
\includegraphics[width=\linewidth]{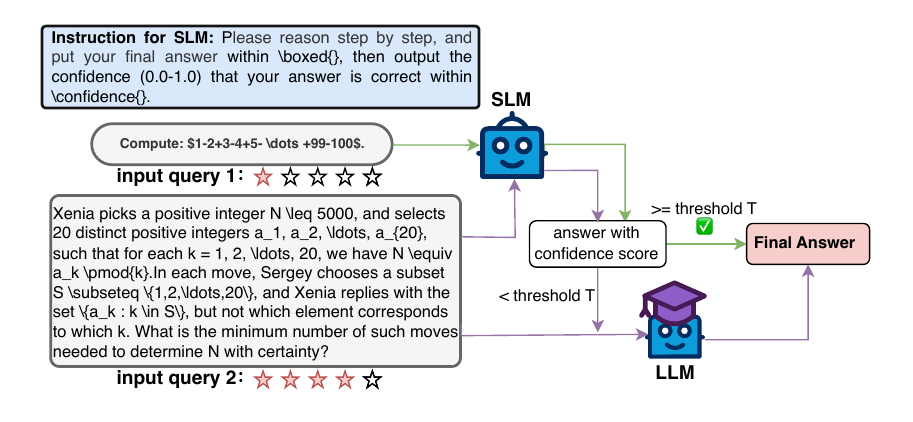}
\caption{Architecture of \model}
\label{fig:architecture}
\end{subfigure}
\hfill
\begin{subfigure}[t]{0.42\textwidth}
\centering
\includegraphics[width=\linewidth]{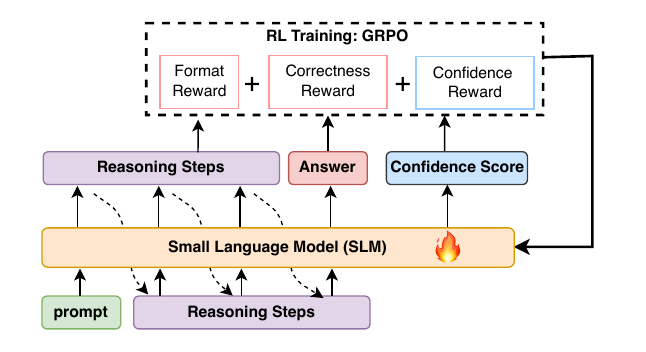}
\caption{Confidence calibration of the SLM}
\label{fig:training_phase}
\end{subfigure}
\caption{
(a) \textbf{Architecture of \textsc{\model}:} \textsc{\model} cascades an SLM with an LLM. For each query, the SLM is prompted to generate reasoning steps, an answer, and a confidence score. LLM is invoked only when the SLM's confidence is less than the predefined threshold.
(b) \textbf{Confidence calibration of the SLM:} The SLM is trained using GRPO with a multi-objective reward function to improve its reasoning ability and calibrate its confidence. }
\label{fig:overall_architecture}
\end{figure*}

In this section, we present the design of our~\model framework. 
We begin by describing the overall system architecture and then detail RL with confidence calibration (RLCC).

\subsection{Overall Architecture: The SLM-LLM Collaborative Framework} 
\label{sec:overall_architecture}

As illustrated in Figure~\ref{fig:architecture}, \model employs a cascaded SLM-LLM architecture where queries are first processed by the SLM. 
The SLM is prompted to generate reasoning steps, an answer, and a verbalized confidence score indicating the likelihood that its answer is correct. 
If the confidence exceeds a predefined threshold $T$, the SLM's answer becomes the final output. Otherwise, the original query is passed to the LLM, whose response serves as the final answer.

To elicit the desired response format, we append the instruction after each question: \textbf{\textit{"Please reason step by step, and put your final answer within \textbackslash{}boxed\{\}, then output the confidence (0.0-1.0) that your answer is correct within \textbackslash{}confidence\{\}."}} 


\subsection{Calibrating Confidence through RL} \label{sec:slm_self_awareness}

RL with verifiable rewards (RLVR) focuses only on improving a model's reasoning ability. RLCC further aligns the model's confidence by adding additional confidence calibration rewards (abbreviated as confidence reward) on top of the verifiable rewards. Moreover, we incorporate a format reward to ensure the model produces well-structured outputs containing reasoning steps, a final answer, and a confidence score.

Formally, given an input question $x$, the SLM is prompted to generate reasoning steps $y_r$, an answer $y_a$, and a confidence score $y_c \in [0, 1]$ using the instruction in Section \ref{sec:overall_architecture}. Let $y^*_a$ denote the ground truth answer and $p$ represent the probability that the SLM can answer the question correctly. The composite reward function is:
\begin{equation}
    \label{eq:final_reward}
    R = R_{\text{correct}} + R_{\text{format}} + R_{\text{confidence}},
\end{equation}
where $R_{\text{correct}}=\mathbb{I}(y_a=y^*_a)$ is the binary correctness reward. $R_{\text{format}}$ is a simple format reward which enforces consistent language usage, output of answer and confidence score as required in the prompt, and is normalized to $[0, 1]$. 

The confidence reward $R_{\text{confidence}}$ should encourage the model to generate confidence scores that aligns with its correctness probability. It is defined as the negative distance between the confidence score $y_c$ and the correctness probability $p$. Several distance metrics can be employed, including L1 distance, L2 distance, and KL distance. Correspondingly, the confidence rewards are
\begin{equation}
\label{eq:l1_confidence_reward}
R_{\text{L1}}(p, y_c) = -|p - y_c|
\end{equation}
\begin{equation}
\label{eq:l2_confidence_reward}
R_{\text{L2}}(p, y_c) = -(p - y_c)^2
\end{equation}
\begin{equation}
\label{eq:kl_confidence_reward}
R_{\text{KL}}(p, y_c) = -p \log\frac{p}{y_c} - (1-p) \log\frac{1-p}{1-y_c}
\end{equation}
$R_{\text{KL}}$ is unbounded, and can lead to instability in training, so we make some clipping, normalization, and remove the constant term to obtain:
\begin{align}
\label{eq:kl_revised_confidence_reward}
R_{\text{KL}}(p, y_c) = & p \frac{\log(\max(y_c, \epsilon))}{\log\epsilon} + \nonumber\\
&(1-p) \frac{\log(\max(1-y_c,\epsilon))}{\log\epsilon},
\end{align}
where $\epsilon\in(0,1)$ is a clipping factor. 

Since the correctness probability $p$ is unavailable, we estimate it using group accuracy during GRPO rollouts. Specifically, for a question-answer pair $(x, y_a^*)\sim\mathcal{D}$, $N$ responses $(y_r^0, y_a^0, y_c^0),\ldots, (y_r^{N-1}, y_a^{N-1}, y_c^{N-1})$ are sampled from the SLM, then $p$ is estimated as:
\begin{equation}
\hat{p} = \frac{1}{N}\sum_{i=0}^{N-1}\mathbb{I}(y_a^i=y_a^*).
\end{equation}

Our confidence reward definition is on group level, i.e., each sampled response $(y_r^i, y_a^i, y_c^i)$ in the same group would use the same $\hat{p}$ regardless of its correctness. This is different from the sample level definitions in \cite{DBLP:journals/corr/abs-2507-16806, DBLP:conf/emnlp/XuWDL0CG24, stangel2025rewardingdoubtreinforcementlearning}. In their definitions, $p$ is estimated as $\hat{p}=\mathbb{I}(y_a^i=y_a^*)$, which depends on the sampled answer's correctness. We denote the sample-level confidence rewards as sample\_L1, sample\_L2 (equivalent to the Brier score in \cite{DBLP:journals/corr/abs-2507-16806}), sample\_KL. We focus on L1 reward and will compare the performance of different rewards in the ablation study. 



\section{Experimental Setup}
\label{sec:experimental_setup}

\begin{table*}[htbp]
\centering
\resizebox{\textwidth}{!}{
\begin{tabular}{lccccccccc}
\toprule[1.5pt]
\multirow{2}{*}{\textbf{System}} & \multicolumn{3}{c}{\textbf{DeepMath500}} & \multicolumn{3}{c}{\textbf{OOD Math}} & \multicolumn{3}{c}{\textbf{OOD Non-Math}} \\
\cmidrule(lr){2-4} \cmidrule(lr){5-7} \cmidrule(lr){8-10}
& \textbf{Pass@1\%} ↑ & \textbf{Avg Cost} ↓ & \textbf{LLM\%} & \textbf{Pass@1\%} ↑ & \textbf{Avg Cost} ↓ & \textbf{LLM\%} & \textbf{Pass@1\%} ↑ & \textbf{Avg Cost} ↓ & \textbf{LLM\%} \\
\midrule[1pt]
\multicolumn{10}{l}{\textbf{\textit{Standalone}}} \\
SLM & 42.7 & 4423 & 0.0 & 60.4 & 2622 & 0.0 & 45.2 & 2004 & 0.0 \\
RLVR-SLM & 57.6 & 2511 & 0.0 & 71.8 & 1928 & 0.0 & 62.4 & 1438 & 0.0 \\
Brier-SLM & 55.7 & 3490 & 0.0 & 72.7 & 2322 & 0.0 & 63.5 & 1708 & 0.0 \\
L1-SLM* & 56.6 & 3766 & 0.0 & 72.1 & 2589 & 0.0 & 64.7 & 1902 & 0.0 \\
\cellcolor{gray!15}Baseline LLM & \cellcolor{gray!15}69.0 & \cellcolor{gray!15}14882 & \cellcolor{gray!15}100.0 & \cellcolor{gray!15}79.6 & \cellcolor{gray!15}10164 & \cellcolor{gray!15}100.0 & \cellcolor{gray!15}71.4 & \cellcolor{gray!15}7428 & \cellcolor{gray!15}100.0 \\
\midrule
\multicolumn{10}{l}{\textbf{\textit{Collaborative}}} \\
SLM-Verb & 60.3\,{\scriptsize(--8.6)} & 11594\,{\scriptsize(--22.1\%)} & 43.6 & 75.7\,{\scriptsize(--3.9)} & 7300\,{\scriptsize(--28.2\%)} & 34.9 & 48.8\,{\scriptsize(--22.6)} & 2609\,{\scriptsize(--64.9\%)} & 9.4 \\
RLVR-SLM-Verb & 57.6\,{\scriptsize(--11.4)} & 2672\,{\scriptsize(--82.0\%)} & 1.0 & 72.6\,{\scriptsize(--7.0)} & 2927\,{\scriptsize(--71.2\%)} & 14.0 & 62.7\,{\scriptsize(--8.7)} & 1510\,{\scriptsize(--79.7\%)} & 1.1 \\
RLVR-SLM-AvgProb & 57.6\,{\scriptsize(--11.4)} & 2511\,{\scriptsize(--83.1\%)} & 0.0 & 72.4\,{\scriptsize(--7.1)} & 2206\,{\scriptsize(--78.3\%)} & 4.5 & 62.5\,{\scriptsize(--8.9)} & 1488\,{\scriptsize(--80.0\%)} & 0.9 \\
RLVR-SLM-Probe & 63.3\,{\scriptsize(--5.7)} & 9562\,{\scriptsize(--35.7\%)} & 41.7 & 75.9\,{\scriptsize(--3.7)} & 5874\,{\scriptsize(--42.2\%)} & 25.2 & 67.3\,{\scriptsize(--4.1)} & 4724\,{\scriptsize(--36.4\%)} & 32.7 \\
Router+RLVR-SLM & 61.4\,{\scriptsize(--7.5)} & 7659\,{\scriptsize(--48.5\%)} & 35.6 & 74.4\,{\scriptsize(--5.2)} & 4974\,{\scriptsize(--51.1\%)} & 20.6 & 65.4\,{\scriptsize(--6.0)} & 3781\,{\scriptsize(--49.1\%)} & 34.0 \\
Brier-SLM-Verb & 66.3\,{\scriptsize(--2.7)} & 15254\,{\scriptsize(+2.5\%)} & 72.5 & \textbf{78.8}\,{\scriptsize(--0.8)} & 9264\,{\scriptsize(--8.8\%)} & 57.4 & \textbf{70.3}\,{\scriptsize(--1.2)} & 7188\,{\scriptsize(--3.2\%)} & 78.9 \\
\model* & \textbf{67.5}\,{\scriptsize(--1.5)} & 13882\,{\scriptsize(--6.7\%)} & 59.9 & 77.9\,{\scriptsize(--1.7)} & 7978\,{\scriptsize(--21.5\%)} & 35.5 & 69.7\,{\scriptsize(--1.7)} & 6178\,{\scriptsize(--16.8\%)} & 59.0 \\
\bottomrule[1.5pt]
\end{tabular}
}
\caption{Performance comparison across different systems and datasets. Pass@1, average cost per question (Avg Cost), and LLM usage percentage (LLM\%) are reported. OOD Math includes Math500, GSM8K, and OlympiadBench. OOD Non-Math includes GPQA and CommonsenseQA. In all collaborative systems, the Baseline LLM's Pass@1 serves as the confidence threshold. Values in parentheses indicate the difference from the Baseline LLM.  \model (L1-SLM-Verb) utilizes L1-SLM and verbalized confidence. * indicates our method.}
\label{tab:main_results}
\end{table*}

\subsection{Models and Dataset}

\textbf{SLM:} Our main base SLM is Qwen2.5-7B-Instruct~\cite{qwen2025qwen25technicalreport} (Qwen2.5-7BI). In ablation studies on model backbones, we additionally evaluate Qwen2.5-1.5B-Instruct~\cite{qwen2025qwen25technicalreport} (Qwen2.5-1.5BI) and Llama3.1-8B-Instruct~\cite{grattafiori2024llama3herdmodels} (Llama3.1-8BI). We use instruction-tuned versions for their superior instruction-following capabilities and faster convergence during training.

\textbf{LLM:} We utilize Qwen2.5-32B-Instruct~\cite{qwen2025qwen25technicalreport} (Qwen2.5-32BI) as the base LLM.

\textbf{Training and In-Domain Evaluation Datasets:} 
We construct our training and in-domain evaluation sets by randomly sampling from the DeepMath-103K~\cite{he2025deepmath103klargescalechallengingdecontaminated} dataset, a meticulously curated mathematical dataset containing 103K question-answer pairs of varying difficulty levels, which also excludes common math benchmark datasets. Specifically, we sample 16K problems for training (referred to as DeepMath16K) and an additional 500 problems as in-domain evaluation dataset (referred to as DeepMath500).

\textbf{Out-of-Domain (OOD) Evaluation Datasets:} We select 5 OOD datasets across mathematical and non-mathematical reasoning tasks. For mathematical reasoning, we utilize Math500~\cite{hendrycks2021measuringmathematicalproblemsolving}, GSM8K~\cite{cobbe2021gsm8k}, and OlympiadBench~\cite{he2024olympiadbenchchallengingbenchmarkpromoting}. For broader reasoning evaluation, we include: 1) GPQA~\cite{rein2024gpqa}, a challenging graduate-level science dataset spanning biology, physics, and chemistry with 448 multiple-choice questions; 2) CommonsenseQA~\cite{talmor-etal-2019-commonsenseqa}, a commonsense reasoning dataset containing 1,221 multiple-choice questions. We evaluate correctness using math\_verify~\cite{mathverify} for mathematical datasets and exact match for others.

\subsection{Baselines}
We consider the following baselines:

\noindent\textbf{Standalone Model:}

\textbullet\ \textbf{SLM:} Original SLM.

\textbullet\ \textbf{RLVR-SLM:} SLM trained using RLVR.

\textbullet\ \textbf{Brier-SLM:} SLM trained using RLCC with the Brier score.

\textbullet\ \textbf{L1-SLM:} SLM trained using RLCC with the L1 confidence reward.

\textbullet\ \textbf{Baseline LLM:}  To make a collaborative system useful, the Baseline LLM should have better reasoning ability than the RL-trained SLM, otherwise, there is no need to defer to the LLM. To meet the requirement, we trained the LLM using RLCC with the L1 confidence reward.  

\noindent\textbf{Collaborative System:} For all systems, we employ the Baseline LLM as the capable LLM. Then the systems differ only in the SLM and the method to generate confidence scores. For simplicity, we use these two factors to denote a system.

\textbullet\ \textbf{SLM-Verb:} The original SLM with verbalized confidence.

\textbullet\ \textbf{RLVR-SLM-Verb:} The RLVR-SLM with verbalized confidence. 

\textbullet\ \textbf{RLVR-SLM-AvgProb:} The RLVR-SLM with confidence score obtained by averaging the token probability of the response. 

\textbullet\ \textbf{RLVR-SLM-Probe:} The RLVR-SLM with a probe to generate confidence score. The probe is a multi-layer perceptron (MLP) which takes as input the average-pooled hidden states from RLVR-SLM's final layer, and outputs the correctness prediction of RLVR-SLM's answer. It is trained using the same DeepMath16K, but with answers sampled from RLVR-SLM. Probe model and training details are provided in Appendix \ref{sec:training_details_probe}.

\textbullet\ \textbf{Router + RLVR-SLM:} We employ a router to decide the use of the SLM or the LLM. We assess the RLVR-SLM's accuracy on DeepMath16k by sampling 8 responses per question, then utilize the accuracy to construct a training dataset for the router with the prompt: \textbf{\textit{"Question: \{Question\} Instruction: Estimate and output the probability (0.0-1.0) that a small language model ($\leq$10B) can answer the question correctly."}} During evaluation, the router generates a probability score for each question. Questions are routed to the RLVR-SLM when the probability exceeds a given threshold, and to the LLM otherwise. The Router utilizes the same base model as RLVR-SLM. Training details are provided in Appendix \ref{sec:training_details_router}.

\textbullet\ \textbf{Brier-SLM-Verb:} The Brier-SLM with verbalized confidence.

\subsection{Evaluation Protocol} The following metrics are considered to evaluate the different systems.

\noindent\textbf{For the SLM component:} 

\textbf{Pass@1:} A common measure of accuracy.

\textbf{Expected Calibration Error (ECE):} Measures the alignment between predicted confidence and actual accuracy through binning:
\begin{equation}
    \text{ECE} = \sum_{m=1}^M \frac{|B_m|}{N} \left|\text{acc}(B_m) - \text{conf}(B_m)\right|,
\end{equation}
where $B_m$ represents the $m$-th confidence bin, $\text{acc}(B_m)$ denotes the empirical accuracy of samples in $B_m$, and $\text{conf}(B_m)$ corresponds to the mean confidence score of predictions in $B_m$.

\textbf{AUROC:} Computed as the area under the ROC curve generated by varying the confidence threshold. For binary classification of answer correctness:
\begin{equation}
    \text{AUROC} = \int_{0}^{1} \text{TPR}(t) \cdot \text{FPR}'(t) dt,
\end{equation}
where TPR (True Positive Rate) and FPR (False Positive Rate) are functions of threshold $t$.

\textbf{Confidence Output Ratio:} Measures the ratio of responses with a verbalized confidence score.

\noindent\textbf{For the collaborative system:} 

\textbf{Pass@1:} Accuracy of the final answer.

\textbf{System Cost:} The system cost depends on the number of input and output tokens. We assume that 1) the cost of output tokens is four times that of input tokens, which aligns with prevailing API pricing models (e.g., GPT 4:1, while Claude 5:1). 2) pricing scales proportionally with model size. For a question, the SLM/Router cost is defined as:
\begin{equation}
\label{eq:slm_cost}
\text{SLM/Router Cost} = n_{\text{prompt}} + 4 \cdot n_{\text{SLM}},
\end{equation}
where $n_{\text{prompt}}$ and $n_{\text{SLM}}$ represent the numbers of prompt and output tokens, respectively. Each SLM input token is normalized to a base cost of 1.

The LLM cost is defined as:
\begin{equation}
\label{eq:llm_cost}
\text{LLM Cost} = \left( n_{\text{prompt}} + 4 \cdot n_{\text{LLM}} \right) \cdot \frac{|\theta_{\text{LLM}}|}{|\theta_{\text{SLM}}|},
\end{equation}
where $n_{\text{LLM}}$ represents the number of output tokens of the LLM, and the term $\frac{|\theta_{\text{LLM}}|}{|\theta_{\text{SLM}}|}$ adjusts for the relative cost scaling based on parameter counts.

The cost of a system without a router is:
$$
{\small
\text{System Cost} = \begin{cases}
\text{SLM Cost}, & y_c \geq p_T \\
\text{SLM Cost} + \text{LLM Cost}, & y_c < p_T
\end{cases}
}
$$
where $y_c$ and $p_T$ are the confidence score of the SLM and the confidence threshold, respectively.

The system cost of Router + RLVR-SLM is:
$$
{\small
\text{System Cost} = \begin{cases}
\text{Router Cost} + \text{SLM Cost}, & y_c \geq p_T \\
\text{Router Cost} + \text{LLM Cost}, & y_c < p_T
\end{cases}
}
$$
where $y_c$ is the correctness probability of the router.

\subsection{Implementation Details}

\begin{figure*}[htbp]
\centering
\includegraphics[width=\linewidth]{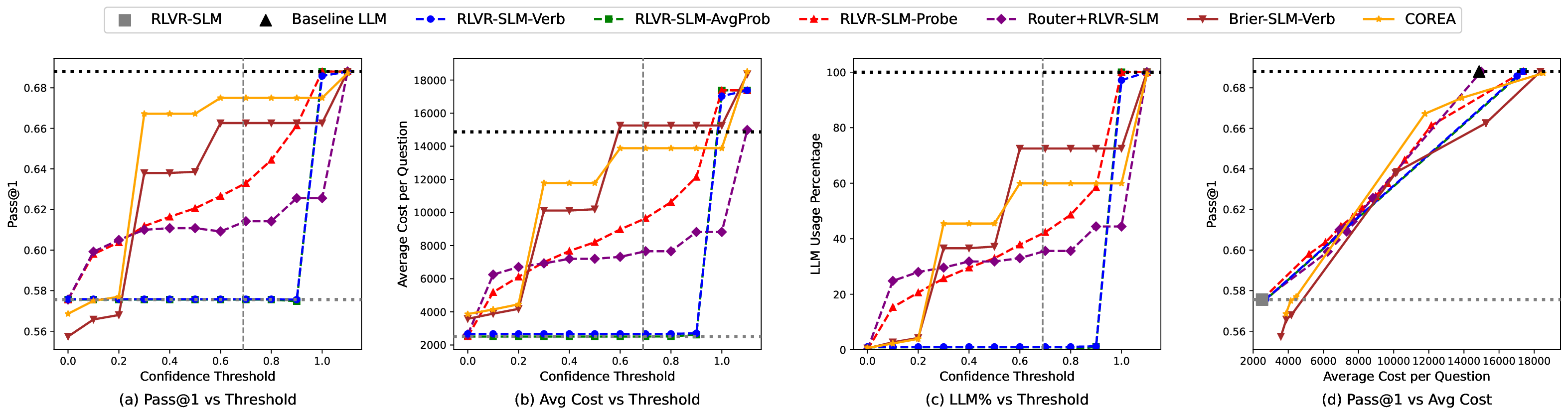}
\caption{Results on DeepMath-500: (a) Pass@1, (b) average cost, and (c) LLM usage percentage versus confidence threshold; (d) Pass@1 versus average cost for different systems. The vertical dashed lines in (a)-(c) mark the Baseline LLM's Pass@1 (0.69).}
\label{fig:performance_theshold}
\end{figure*}

We conduct training on an AWS p5en.48xlarge instance equipped with 8×H200 GPUs using the SimpleRL-reason framework~\cite{zeng2025simplerlzooinvestigatingtamingzero}. 
For GRPO training, we use the framework's default hyperparameters with the following key settings: sampling temperature of 1.0, maximum response length of 4,096 tokens, and a learning rate of $5 \times 10^{-7}$ with AdamW optimizer. 
Each training episode consists of two phases: (1) a rollout phase generating 8 responses per question with a batch size of 1,024, and (2) a training phase using a batch size of 256. 
The loss function includes KL divergence regularization (coefficient: $1 \times 10^{-4}$) to prevent policy drift and entropy regularization (coefficient: $1 \times 10^{-3}$) to encourage exploration. We trained Qwen2.5-1.5BI, Qwen2.5-7BI, Llama3.1-8BI, and Qwen2.5-32BI for 150, 80, 120, and 50 steps, respectively, with training duration determined by total reward convergence. During inference, we set the sampling temperature to 0.6 to reduce output variability and improve consistency. All evaluation results are averaged over 10 runs unless otherwise stated.

\section{Results}

\subsection{Overall Performance}

The main results are provided in Table \ref{tab:main_results}. Detailed per-dataset results and experiments with an alternative SLM+LLM pairing are provided in Appendix~\ref{subsec:main_results_dataset} and Appendix~\ref{sec:slm1.5b_llm32b}, respectively. From the table, we observe that:

\textbf{Confidence reward does not hurt correctness}: This is demonstrated by comparing the standalone models. Compared to RLVR-SLM, both Brier-SLM and L1-SLM achieve comparable Pass@1 performance despite incorporating an additional confidence reward. Furthermore, although all RL trainings are conducted on DeepMath16K, performance improvements extend to the OOD Math and OOD Non-Math datasets as well, demonstrating the good generalizability of the proposed method.

\textbf{\model shows superior performance-cost trade-off:} In Table \ref{tab:main_results}, we set the Baseline LLM's Pass@1 as the confidence threshold for all collaborative systems, where the SLM defers to the LLM when its confidence falls below this threshold. With well-calibrated confidence, the system should achieve substantial cost reductions while maintaining performance, as confident questions are answered by the SLM alone. Our \model achieves this goal, reducing costs by 6.7\%, 21.5\%, and 16.8\% on DeepMath500, OOD Math, and OOD Non-Math respectively, while limiting Pass@1 degradation to within 2 percentage points.

While some systems achieve either higher Pass@1 or lower costs in certain cases—for instance, Brier-SLM-Verb attains higher Pass@1 on OOD datasets—this comes at the expense of reduced savings. Conversely, systems like RLVR-SLM-Verb, RLVR-SLM-AvgProb, RLVR-SLM-Probe, and Router-RLVR-SLM achieve greater cost reductions but suffer significant Pass@1 drops due to poorly calibrated confidence scores. These results demonstrate that our method successfully maintains near-optimal performance while achieving substantial cost savings through intelligent SLM-LLM collaboration.

\subsection{Confidence Threshold Selection}

In Table \ref{tab:main_results}, the confidence threshold is fixed at the Baseline LLM's Pass@1. However, the collaborative system offers flexibility to trade off performance and cost by adjusting this threshold. As shown in Figure~\ref{fig:performance_theshold}, when the threshold increases from 0 to 1.1, system Pass@1 increases alongside average cost due to higher LLM usage across all methods. If cost is prioritized, we can sacrifice accuracy for savings. For instance, decreasing the confidence threshold from 0.69 to 0.4 for L1-SLM-Verb yields an additional 14\% relative cost reduction with only 0.6 percentage point Pass@1 loss. While other methods achieve greater cost reductions, they incur more severe accuracy drops. We set 1.1 as the maximum threshold since the SLM's verbalized confidence caps at 1.0, thus only values > 1.0 ensure all questions defer to the LLM. Figure~\ref{fig:performance_theshold}d draws Pass@1 versus average cost curves of different methods, although \model is not Pareto optimal across all regions, it achieves the best accuracy-cost trade-off in high Pass@1 regions ($[0.62, 0.67]$).

\subsection{Evolution of SLM Self-Awareness} 

\begin{figure*}[htbp]
\centering
\begin{subfigure}[t]{0.98\textwidth}
\centering
\includegraphics[width=\linewidth]{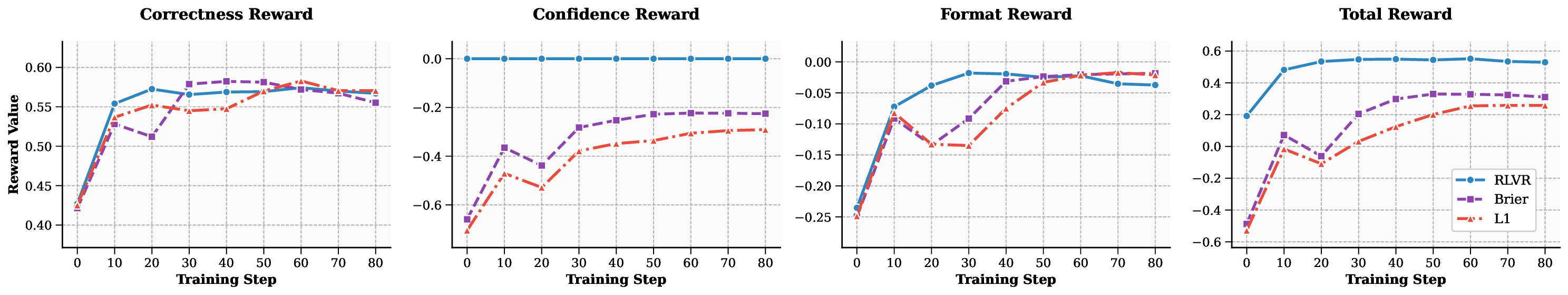}
\caption{Reward evolution.}
\label{fig:reward_evolution}
\end{subfigure}
\hfill
\begin{subfigure}[t]{0.98\textwidth}
\centering
\includegraphics[width=\linewidth]{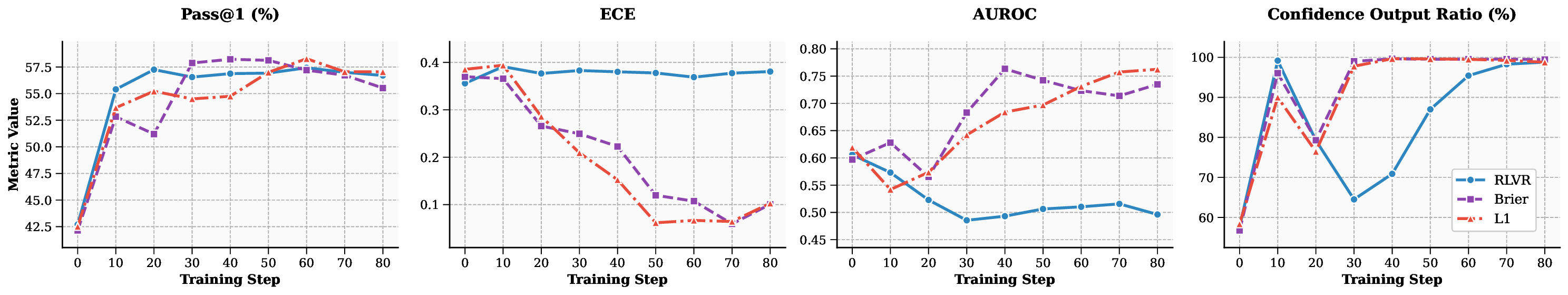}
\caption{Pass@1, ECE, AUROC, Confidence Output Ratio evolution.}
\label{fig:metric_evolution}
\end{subfigure}
\begin{subfigure}[t]{0.98\textwidth}
\centering
\includegraphics[width=\linewidth]{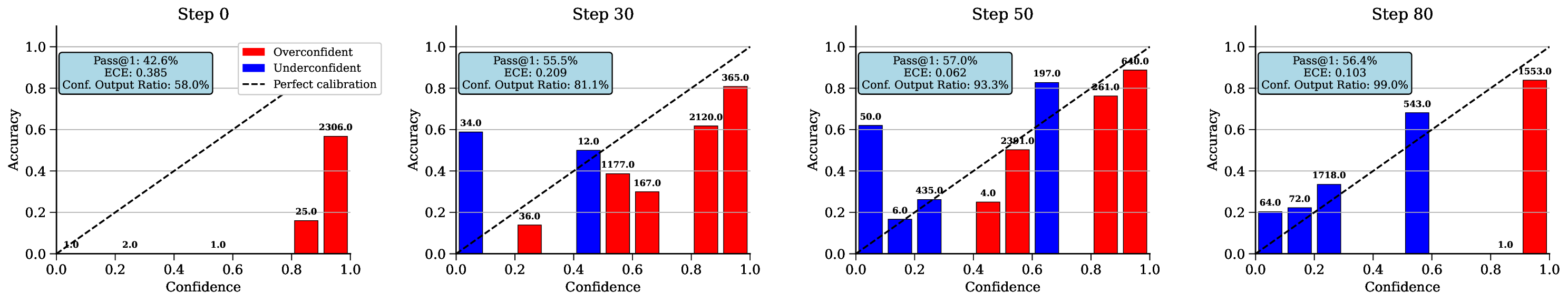}
\caption{Confidence alignment evolution of L1, the number on each bar indicates how many answers are in the confidence bin.}
\label{fig:conf_acc_evolution}
\end{subfigure}
\caption{Training process of Qwen2.5-7BI with different reward configurations: RLVR, Brier (RLCC with Brier score), and L1 (RLCC with L1 confidence reward) on DeepMath500, with 8 sampled answers per question.}
\label{fig:training_process}
\end{figure*}

\begin{table}[t]
\centering
\resizebox{0.98\columnwidth}{!}{%
\begin{tabular}{l c c c}
\toprule
\textbf{Method} & \textbf{Pass@1\%} ↑ & \textbf{ECE} ↓ & \textbf{AUROC} ↑ \\
\midrule
\cellcolor{gray!15}SLM-Verb & \cellcolor{gray!15}$52.8 \pm 0.9$ & \cellcolor{gray!15}$0.45 \pm 0.01$ \cellcolor{gray!15}& \cellcolor{gray!15}$0.50 \pm 0.01$ \\
RLVR-SLM-Verb & $66.9 \pm 0.7$ & $0.30 \pm 0.01$ & $0.50 \pm 0.01$ \\
RLVR-SLM-AvgProb & $66.9 \pm 0.7$ & $0.28 \pm 0.01$ & $0.64 \pm 0.01$ \\
RLVR-SLM-Probe & $66.9 \pm 0.7$ & $0.17 \pm 0.01$ & $\mathbf{0.72 \pm 0.01}$ \\
Router+RLVR-SLM & $66.9 \pm 0.7$ & $0.20 \pm 0.01$ & $0.56 \pm 0.01$ \\
Brier-SLM-Verb & $67.5 \pm 0.7$ & $0.17 \pm 0.01$ & $\mathbf{0.72 \pm 0.01}$ \\
L1-SLM-Verb* & $\mathbf{67.8 \pm 0.7}$ & $\mathbf{0.12 \pm 0.01}$ & $\mathbf{0.72 \pm 0.01}$ \\
\bottomrule
\end{tabular}
}
\caption{Comparison of different methods on accuracy and calibration metrics, averaged across all datasets. Here, the metrics are only evaluated on the SLM.}
\label{tab:slm_calibration_results}
\end{table}

\textbf{Training Dynamics and Convergence:} Figure \ref{fig:reward_evolution} demonstrates that trainings with and without confidence rewards (either Brier Score or L1 confidence reward) achieve comparable correctness and format rewards, indicating that the SLM can simultaneously improve its reasoning ability and align its confidence estimates. The evolution of key metrics in Figure \ref{fig:metric_evolution} (Pass@1, ECE, AUROC, and Confidence Output Ratio) further supports this finding. Figure \ref{fig:conf_acc_evolution} provides a clearer view of the confidence-accuracy alignment process: initially (step 0), the SLM exhibits severe overconfidence with nearly all confidence scores approaching 1.0. Through training, the model progressively diversifies its confidence estimates and aligns them more closely with actual accuracies.

\textbf{Quantitative Results:} Table \ref{tab:slm_calibration_results} shows the averaged accuracy and calibration results of different methods on all datasets (detailed results on each dataset are presented in Appendix \ref{sec:detailed_calibration}). It is noted that RL training improves the SLM's reasoning performance significantly. Among all methods, the proposed L1-SLM-Verb method achieves the highest Pass@1 and lowest ECE, showing the superiority of the proposed method. 

\subsection{Ablation Study}
\subsubsection{Generalizability: Robustness Across SLM Sizes and Architectures}

\begin{figure}[h]
    \centering
    \includegraphics[width=\linewidth]{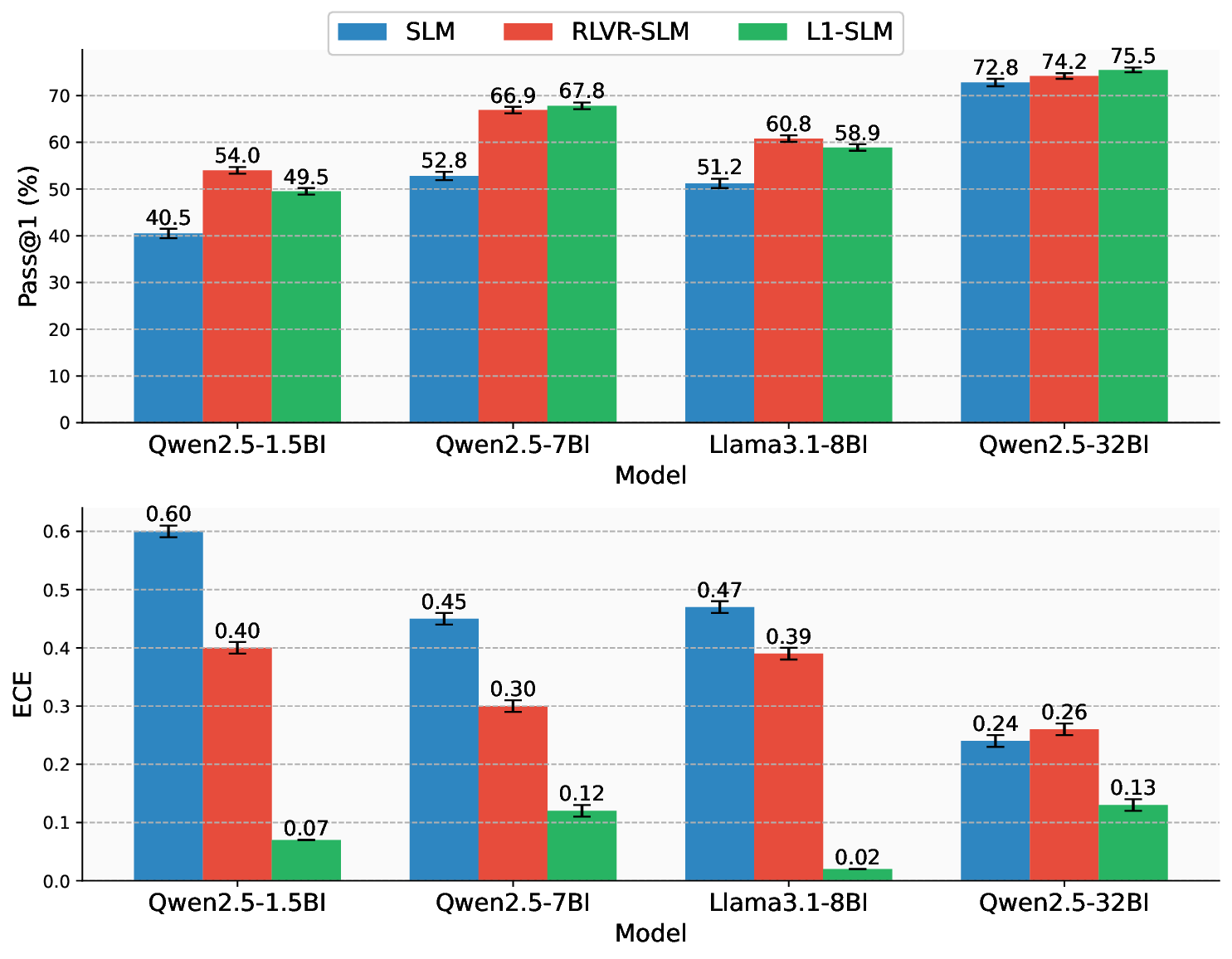}
    \caption{Comparison of Pass@1 and ECE across several models trained with different reward configurations, averaged across all datasets. Detailed results on each dataset are provided in Appendix \ref{sec:detailed_calibration}.}
    \label{fig:model_comparison}
\end{figure}

We further conducted RLVR training and RLCC training with L1 confidence reward (RLCC-L1) on Qwen2.5-1.5B, Llama3.1-8B, and Qwen2.5-32B. Figure \ref{fig:model_comparison} demonstrates that the proposed RLCC method is effective across all tested model sizes and architectures. In terms of reasoning performance, both RLVR and RLCC-L1 substantially improve capabilities for smaller models, with the exception of Qwen2.5-32B, which already exhibits strong baseline performance. For calibration, RLCC-L1 consistently reduces ECE by a large margin across all models, while RLVR yields only slight ECE reductions or even degradations.

\subsubsection{Different Reward Functions}
\label{sec:results_different_confs}

\begin{table}[t]
\centering
\resizebox{0.98\columnwidth}{!}{%
\begin{tabular}{l c c c}
\toprule
\textbf{Conf. Reward} & \textbf{Pass@1\%} $\uparrow$ & \textbf{ECE} $\downarrow$ & \textbf{AUROC} $\uparrow$ \\
\midrule
\cellcolor{gray!15}L1 & \cellcolor{gray!15}$67.8 \pm 0.7$ & \cellcolor{gray!15}$0.12 \pm 0.01$ & \cellcolor{gray!15}$0.72 \pm 0.01$ \\
Sample\_L1 & $\mathbf{68.3 \pm 0.7}$ & $0.16 \pm 0.01$ & $0.68 \pm 0.01$ \\
L2 & $67.5 \pm 0.7$ & $0.31 \pm 0.01$ & $0.62 \pm 0.01$ \\
\cellcolor{gray!15}Brier & \cellcolor{gray!15}$67.5 \pm 0.7$ & \cellcolor{gray!15}$0.17 \pm 0.01$ & \cellcolor{gray!15}$0.72 \pm 0.01$ \\
KL & $66.5 \pm 0.7$ & $0.15 \pm 0.01$ & $0.69 \pm 0.01$ \\
Sample\_KL & $67.1 \pm 0.7$ & $\mathbf{0.11 \pm 0.01}$ & $\mathbf{0.74 \pm 0.01}$ \\
\bottomrule
\end{tabular}
}
\caption{Comparison of different confidence reward functions on accuracy and calibration metrics, averaged across all datasets.}
\label{tab:results_different_confs}
\end{table}

We compare the performance of RLCC with different confidence rewards in Table~\ref{tab:results_different_confs}. It can be concluded that RLCC training with these confidence rewards achieve comparable performances in terms of accuracy and confidence calibration, except for the L2 reward, which has a much higher ECE than the others. We hypothesize that this phenomenon stems from instabilities during training. Specifically, although the overall reward plateaus, the model continues to adjust among different rewards. As we adopt a fixed stopping step across all configurations, the selected checkpoint may not represent the optimal point for the L2 reward. Interestingly, even Sample\_L1 is not a proper function \cite{DBLP:journals/corr/abs-2507-16806} (which means that theoretically the highest reward is not attained at $y_c = p$ in Equation (\ref{eq:l1_confidence_reward})), training with it can still achieve similar performance as other proper functions. Among all, we recommend L1 reward as it achieves a better balance between Pass@1 and ECE.

\subsubsection{Impact of Rollout Size}

\begin{table}[t]
\centering
\resizebox{0.98\columnwidth}{!}{%
\begin{tabular}{c c c c}
\toprule
\textbf{Rollout Size} & \textbf{Pass@1\%} $\uparrow$ & \textbf{ECE} $\downarrow$ & \textbf{AUROC} $\uparrow$ \\
\midrule
4 & $67.2 \pm 0.8$ & $\mathbf{0.10 \pm 0.01}$ & $0.68 \pm 0.01$ \\
\cellcolor{gray!15}8 & \cellcolor{gray!15}$67.8 \pm 0.7$ & \cellcolor{gray!15}$0.12 \pm 0.01$ & \cellcolor{gray!15}$\mathbf{0.72 \pm 0.01}$ \\
16 & $66.7 \pm 0.7$ & $\mathbf{0.10 \pm 0.01}$ & $0.69 \pm 0.01$ \\
32 & $\mathbf{68.3 \pm 0.5}$ & $0.17 \pm 0.01$ & $0.65 \pm 0.01$ \\
\bottomrule
\end{tabular}
}
\caption{Comparison of different rollout sizes with L1 confidence reward on accuracy and calibration metrics, averaged across all datasets.}
\label{tab:rollout}
\end{table}

Table~\ref{tab:rollout} presents the performance of different rollout sizes in GRPO training with L1 confidence reward. Contrary to the expectation that larger rollout sizes would monotonically improve both accuracy and calibration through increased exploration diversity and better probability estimation, the results reveal a more nuanced relationship. Rollout size 32 achieves the highest Pass@1 (68.3\%) but exhibits degraded calibration metrics (ECE: 0.17, AUROC: 0.65). Conversely, rollout size 16 attains the best calibration metrics (ECE: 0.10, AUROC: 0.69), yet suffers a slight drop in Pass@1 to 66.7\%. Rollout size 8 demonstrates a favorable balance, achieving competitive Pass@1 while maintaining strong calibration. This indicates that the proposed method is not sensitive to rollout size, even with a small rollout size like 4 could perform well.

\section{Conclusion}
\label{sec:conclusion}

We introduced \textbf{\model}, a collaborative reasoning framework in which an SLM is trained with RL to calibrate its confidence estimates and strategically defer challenging questions to an LLM when uncertain. Through extensive experiments, we demonstrated that the proposed RLCC training method simultaneously enhances the SLM's reasoning capabilities and aligns its confidence scores across multiple SLM backbones and model scales. These improvements generalize effectively to both mathematical and non-mathematical reasoning tasks. Furthermore, \model demonstrated superior performance-cost tradeoffs compared to strong baselines. By intelligently deferring questions to the LLM based on calibrated confidence scores, our approach achieved substantial cost reductions of $7\%\sim22\%$ while maintaining accuracy within 2 percentage points of the baseline LLM. These results highlight that well-calibrated confidence serves as a powerful mechanism for enabling practical and efficient SLM-LLM collaboration in real-world deployment scenarios.

\section*{Limitations}

While the proposed framework shows promising results, several limitations remain. First, the verbalized confidence score after training still tend to be discrete instead of continuous although the prompt instructs the model to output a continuous value in $[0,1]$. This leads to a non-smooth transition of accuracy and cost when we move the confidence threshold. Second, in rare cases, the RL training process can still encounter instabilities, and the model can still balance among different reward components even when the composite reward converges. Hence, how to ensure all components simultaneously converge is worth further study. Finally, we did not extensively discuss different SLM and LLM combinations although we provided a different SLM-LLM pairing in the appendix, however, the parameter count disparity between the SLM and LLM has a significant impact on the cost gain. This also hints that the cost reduction estimation in this paper can be conservative, if LLM is significantly larger than the SLM, the possible gain could be much larger. These factors constrain the current scope of the study and point to directions for further investigation.


\bibliography{eacl2026}

\appendix
\clearpage
\section*{Appendix}

\section{Training Details}
\label{sec:training_details}

\subsection{Training Details of Probe} \label{sec:training_details_probe}

\textbf{Model Architecture.} The Probe takes the average-pooled hidden states from the last layer of the SLM as input and outputs the probability that the SLM's answer is correct. We use a 2-layer MLP with dimensions (2048, 1024) for the two layers. The final head is a sigmoid function.

\textbf{Training Dataset.} The Probe is trained to predict the RLVR-SLM's correctness. We first use RLVR-SLM to generate a response for each question in DeepMath16K, then evaluate the correctness of each response as the ground truth. Specifically, each item in the training dataset is formatted as ((question, answer), correctness). We use a 9:1 train-eval split.

\textbf{Training Details.} We conduct training on an AWS p5en.48xlarge instance equipped with 8×H200 GPUs. We use a per-device batch size of 32 and a learning rate of $5 \times 10^{-5}$ with the AdamW optimizer. We train for 20 epochs until the loss plateaus on the evaluation dataset.

\subsection{Training Details of Router} \label{sec:training_details_router}

\textbf{Model Architecture.} The Router is used to decide whether to allocate a question to the SLM or the LLM. It evaluates the probability that the SLM can answer the question correctly. To achieve this, we use the same base model as the SLM to serve as the Router. We find that directly prompting the SLM to make the prediction is highly inaccurate, so we use supervised fine-tuning to align the Router's predictions.

\textbf{Training Dataset.} The Router is trained to estimate the RLVR-SLM's accuracy on each question. We first use RLVR-SLM to generate 8 responses for each question in DeepMath16K, then compute the accuracy as the ground truth for the question. The prompt is: \textit{"Question: \{Question\} Instruction: Estimate and output the probability (0.0-1.0) that a small language model ($\leq$10B) can answer the question correctly."} We then convert the dataset into the ShareGPT format.

\textbf{Training Details.} We conduct training on an AWS p5en.48xlarge instance. We use LLaMA-Factory~\cite{zheng2024llamafactory} as the training framework. The per-device batch size is 8, gradient accumulation steps is 4, and the learning rate is $5 \times 10^{-5}$ with the AdamW optimizer. We train for 2 epochs until the loss plateaus on the evaluation dataset.

\section{Sample Output}

We use the following model abbreviations throughout this paper:
The "I" suffix indicates instruction-tuned model variants (e.g., 7BI = 7B-Instruct). 
\begin{itemize}
    \item \textbf{Qwen2.5-7BI}: Original Qwen2.5-7B-Instruct model without additional training. 
    \item \textbf{RLVR-Qwen2.5-7BI}: Qwen2.5-7B-Instruct trained with reinforcement learning with verifiable reward (RLVR). 
    \item \textbf{L1-Qwen2.5-7BI}: Qwen2.5-7B-Instruct trained with RL with an additional L1 confidence reward. 
\end{itemize}

Figures~\ref{fig:SLM_output}--\ref{fig:L1-SLM_output} show the responses of Qwen2.5-7BI, RLVR-Qwen2.5-7BI, and L1-Qwen2.5-7BI on a DeepMath500 problem, respectively. As demonstrated, Qwen2.5-7BI produces an incorrect answer, while both RLVR-Qwen2.5-7BI and L1-Qwen2.5-7BI solve the problem correctly, demonstrating the effectiveness of RL training in improving the model's reasoning ability.

Additionally, Qwen2.5-7BI fails to follow the instruction to output the confidence score in the required format. In contrast, both RLVR-Qwen2.5-7BI and L1-Qwen2.5-7BI successfully output scores in the correct format. However, without confidence calibration, RLVR-Qwen2.5-7BI produces an overconfident score of 1.0, similar to Qwen2.5-7BI. Conversely, L1-Qwen2.5-7BI's confidence score of 0.8 is better aligned with the model's true accuracy of 0.75, demonstrating the effectiveness of the confidence reward term in improving confidence calibration.

\begin{figure*}[!t]
\begin{tcolorbox}[
    colback=lightgray,
    colframe=darkgray,
    coltitle=white,
    title={\textbf{Question}},
    arc=5pt,        
    rounded corners,
    boxrule=2pt
]

Given that \( a > b > c > d \), find the maximum integer value of \( n \) such that \[ \frac{1}{a-b} + \frac{1}{b-c} + \frac{1}{c-d} \geq \frac{n}{a-d}. \]

Please reason step by step, and put your final answer within \textbackslash boxed\{\}, then output the confidence ($0.0\sim1.0$) that your answer is correct within \textbackslash confidence\{\}.

(Ground Truth Answer: $9$)

\end{tcolorbox}

\begin{tcolorbox}[
    colback=lightblue,
    colframe=darkblue,
    coltitle=white,
    title={\textbf{Qwen2.5-7BI}},
    arc=5pt,    
    rounded corners,
    boxrule=2pt
]
To find the maximum integer value of \( n \) such that the inequality \[ \frac{1}{a-b} + \frac{1}{b-c} + \frac{1}{c-d} \geq \frac{n}{a-d} \] holds for all \( a > b > c > d \), we start by considering the properties of the terms involved. Let's denote \( x = a - d \), \( y = a - c \), \( z = a - b \), and \( w = b - c \). Since \( a > b > c > d \), we have \( x > y > z > w > 0 \). The inequality then becomes \[ \frac{1}{z} + \frac{1}{w} + \frac{1}{z-w} \geq \frac{n}{x} = \frac{n}{y+z}. \]

To find the maximum possible value of \( n \), we need to analyze the expression on the left side. Let's consider a specific case where \( a - d = 4 \), \( a - c = 3 \), \( a - b = 2 \), and \( b - c = 1 \). In this case, \( x = 4 \), \( y = 3 \), \( z = 2 \), and \( w = 1 \), so the inequality is \[ \frac{1}{2} + \frac{1}{1} + \frac{1}{2-1} \geq \frac{n}{3+2} = \frac{n}{5}. \] Simplifying the left side, we get \[ \frac{1}{2} + 1 + 1 = \frac{1}{2} + 2 = \frac{5}{2}. \] Therefore, the inequality becomes \[ \frac{5}{2} \geq \frac{n}{5}, \] which simplifies to \[ n \leq \frac{25}{2} = 12.5. \] Since \( n \) must be an integer, the maximum possible value of \( n \) is 12.
\end{tcolorbox}
\caption{Question and Response of Qwen2.5-7BI.}
\label{fig:SLM_output}
\end{figure*}

\begin{figure*}[!t]

\begin{tcolorbox}[
    colback=lightblue,
    colframe=darkblue,
    coltitle=white,
    title={\textbf{Qwen2.5-7BI continued}},
    arc=5pt,   
    rounded corners,
    boxrule=2pt
]
To confirm that \( n = 12 \) is indeed the maximum value, we need to check if the inequality holds for \( n = 12 \) in general. We rewrite the inequality as \[ \frac{1}{a-b} + \frac{1}{b-c} + \frac{1}{c-d} \geq \frac{12}{a-d}. \] Using the Cauchy-Schwarz inequality in the form \[ \left( \frac{1}{a-b} + \frac{1}{b-c} + \frac{1}{c-d} \right) \left( (a-b) + (b-c) + (c-d) \right) \geq (1+1+1)^2 = 9, \] we get \[ \left( \frac{1}{a-b} + \frac{1}{b-c} + \frac{1}{c-d} \right) (a-d) \geq 9. \] This simplifies to \[ \frac{1}{a-b} + \frac{1}{b-c} + \frac{1}{c-d} \geq \frac{9}{a-d}. \] Since \(\frac{9}{a-d} < \frac{12}{a-d}\), the inequality \(\frac{1}{a-b} + \frac{1}{b-c} + \frac{1}{c-d} \geq \frac{12}{a-d}\) is indeed satisfied for \( n = 12 \).

To check if \( n = 13 \) works, we consider the case where \( a = 4 \), \( b = 3 \), \( c = 2 \), and \( d = 1 \). Then the inequality becomes \[ \frac{1}{1} + \frac{1}{1} + \frac{1}{1} \geq \frac{13}{3}, \] which simplifies to \[ 3 \geq \frac{13}{3}, \] or \[ 9 \geq 13, \] which is false. Therefore, \( n = 13 \) does not work.

Thus, the maximum integer value of \( n \) is \(\boxed{12}\). The confidence in this answer is 1.0.
\end{tcolorbox}
\caption{Response of Qwen2.5-7BI continued.}
\label{fig:SLM_output_continued}
\end{figure*}

\begin{figure*}[!t]
\begin{tcolorbox}[
    colback=lightgreen,
    colframe=darkgreen,
    coltitle=white,
    title={\textbf{RLVR-Qwen2.5-7BI}},
    arc=5pt,        
    rounded corners,
    boxrule=2pt
]

To find the maximum integer value of $n$ such that the inequality

\[
\frac{1}{a-b} + \frac{1}{b-c} + \frac{1}{c-d} \geq \frac{n}{a-d}
\]

holds for any $a > b > c > d$, we need to analyze the expression and find a suitable value for $n$. We will use the method of substitution and simplification to derive the value of $n$.

First, let's consider the case where $a, b, c,$ and $d$ are equally spaced. This will help us to find a potential value for $n$. If we assume $a = d + 3x$, $b = d + 2x$, and $c = d + x$, then the differences are $a - b = x$, $b - c = x$, and $c - d = x$. Substituting these into the inequality, we get:

\[
\frac{1}{x} + \frac{1}{x} + \frac{1}{x} \geq \frac{n}{3x}.
\]

This simplifies to:

\[
\frac{3}{x} \geq \frac{n}{3x}.
\]

By canceling $x$ from both sides (assuming $x > 0$), we get:

\[
3 \geq \frac{n}{3} \implies n \leq 9.
\]

This tells us that the maximum possible value for $n$ is 9. Now, we need to verify that $n = 9$ is indeed the maximum value that satisfies the inequality for all $a > b > c > d$. To do this, we can use the method of inequalities and specifically the Cauchy-Schwarz inequality in the form of the Titu's lemma (also known as the Engel's form of the Cauchy-Schwarz inequality):

\[
\frac{1^2}{a-b} + \frac{1^2}{b-c} + \frac{1^2}{c-d} \geq \frac{(1+1+1)^2}{(a-b) + (b-c) + (c-d)} = \frac{9}{a-d}.
\]

This inequality shows that the left side is always at least $\frac{9}{a-d}$, which confirms that the maximum value of $n$ is indeed 9.

Therefore, the maximum integer value of $n$ is 9. The confidence in this answer is very high, as it is derived from a general inequality and a specific case analysis.

The final answer is:

\[
\boxed{9}
\]

The confidence in this answer is:

\textbackslash confidence\{1.0\}
\end{tcolorbox}
\caption{Response of RLVR-Qwen2.5-7BI.}
\label{fig:RLVR-SLM_output}
\end{figure*}

\begin{figure*}[!t]

\begin{tcolorbox}[
    colback=lightpurple,
    colframe=purple,
    coltitle=white,
    title={\textbf{L1-Qwen2.5-7BI}},
    arc=5pt,        
    rounded corners,
    boxrule=2pt
]

To solve the inequality \[ \frac{1}{a-b} + \frac{1}{b-c} + \frac{1}{c-d} \geq \frac{n}{a-d}, \] we need to find the maximum integer value of \( n \) such that the inequality holds for all \( a > b > c > d \). 

\#\#\# Step 1: Simplify the Problem
First, let's consider the case when the terms in the inequality are equal. This will help us find an upper bound for \( n \). If we assume \( a-b = b-c = c-d \), then we can simplify the inequality.

\#\#\# Step 2: Assume Equal Differences
Let \( a-b = b-c = c-d = x \). Then, we have:
\[ a = d + 3x. \]

\#\#\# Step 3: Substitute and Simplify
Substitute these values into the inequality:
\[ \frac{1}{x} + \frac{1}{x} + \frac{1}{x} \geq \frac{n}{3x}. \]

This simplifies to:
\[ \frac{3}{x} \geq \frac{n}{3x}. \]

\#\#\# Step 4: Solve for \( n \)
We can cancel \( x \) from both sides of the inequality (since \( x > 0 \)):
\[ 3 \geq \frac{n}{3}. \]

Multiplying both sides by 3 gives:
\[ 9 \geq n. \]

\#\#\# Step 5: Verify the Maximum Value
We need to verify that \( n = 9 \) is indeed the maximum value that satisfies the inequality for all \( a > b > c > d \). To do this, we use the Cauchy-Schwarz inequality in the form of the Titu's lemma (also known as the Engel's form of the Cauchy-Schwarz inequality).

\#\#\# Step 6: Apply Titu's Lemma
Titu's lemma states that for any real numbers \( x_i \) and positive real numbers \( y_i \):
\[ \frac{x_1^2}{y_1} + \frac{x_2^2}{y_2} + \cdots + \frac{x_k^2}{y_k} \geq \frac{(x_1 + x_2 + \cdots + x_k)^2}{y_1 + y_2 + \cdots + y_k}. \]

Applying Titu's lemma to our inequality, we get:
\[ \frac{1}{a-b} + \frac{1}{b-c} + \frac{1}{c-d} \geq \frac{(1+1+1)^2}{(a-b) + (b-c) + (c-d)} = \frac{9}{a-d}. \]

\#\#\# Step 7: Conclusion
From the application of Titu's lemma, we see that the inequality holds with \( n = 9 \). Therefore, the maximum integer value of \( n \) is indeed 9.

\[
\boxed{9}
\]

\textbackslash confidence\{0.8\}

\end{tcolorbox}
\caption{Response of L1-Qwen2.5-7BI (our approach).}
\label{fig:L1-SLM_output}
\end{figure*}

\section{Main Results for Each Dataset}
\label{subsec:main_results_dataset}

\begin{table*}[t]
\centering
\resizebox{0.8\textwidth}{!}{
\begin{tabular}{lccccc}
\toprule
\textbf{System} & \textbf{Pass@1\%} & \textbf{SLM Cost} & \textbf{LLM Cost} & \textbf{Total Cost} & \textbf{LLM\%} \\
\midrule
\multicolumn{6}{l}{\textbf{\textit{Standalone}}} \\
SLM & 42.7 & 4423 & 0 & 4423 & 0 \\
RLVR-SLM & 57.6 & 2511 & 0 & 2511 & 0 \\
Brier-SLM & 55.7 & 3490 & 0 & 3490 & 0 \\
L1-SLM* & 56.6 & 3766 & 0 & 3766 & 0 \\
\cellcolor{gray!15}Baseline LLM & \cellcolor{gray!15}69.0 & \cellcolor{gray!15}0 & \cellcolor{gray!15}14882 & \cellcolor{gray!15}14882 & \cellcolor{gray!15}100 \\
\midrule
\multicolumn{6}{l}{\textbf{\textit{Collaborative}}} \\
SLM-Verb & 60.3 {\scriptsize (-8.6)} & 4423 & 7171 & 11594 {\scriptsize (-22.1\%)} & 43.6 \\
RLVR-SLM-Verb & 57.6 {\scriptsize (-11.4)} & 2511 & 160 & 2672 {\scriptsize (-82.0\%)} & 1.0 \\
RLVR-SLM-AvgProb & 57.6 {\scriptsize (-11.4)} & 2511 & 0 & 2511 {\scriptsize (-83.1\%)} & 0 \\
RLVR-SLM-Probe & 63.3 {\scriptsize (-5.7)} & 2511 & 7051 & 9562 {\scriptsize (-35.7\%)} & 41.7 \\
Router+RLVR-SLM & 61.4 {\scriptsize (-7.5)} & 1503 & 6040 & 7659 {\scriptsize (-48.5\%)} & 35.6 \\
Brier-SLM-Verb & 66.3 {\scriptsize (-2.7)} & 3490 & 11765 & 15254 {\scriptsize (+2.5\%)} & 72.5 \\
\model* & 67.5 {\scriptsize (-1.5)} & 3766 & 10116 & 13882 {\scriptsize (-6.7\%)} & 59.9 \\
\bottomrule
\end{tabular}
}
\caption{Performance comparison across different systems on DeepMath500. Pass@1, SLM Cost per question, LLM Cost per question, Total Cost per question, and LLM usage percentage (LLM\%) are reported. In all collaborative systems, the Baseline LLM's Pass@1 serves as the confidence threshold. Values in parentheses indicate the difference from the Baseline LLM.  \model (L1-SLM-Verb) utilizes L1-SLM and verbalized confidence. * indicates our method.}
\label{tab:main_results_deepmath500}
\end{table*}

\begin{table*}[t]
\centering
\resizebox{0.8\textwidth}{!}{
\begin{tabular}{lccccc}
\toprule
\textbf{System} & \textbf{Pass@1\%} & \textbf{SLM Cost} & \textbf{LLM Cost} & \textbf{Total Cost} & \textbf{LLM\%} \\
\midrule
\multicolumn{6}{l}{\textbf{\textit{Standalone}}} \\
SLM & 54.1 & 3113 & 0 & 3113 & 0 \\
RLVR-SLM & 70.5 & 2125 & 0 & 2125 & 0 \\
Brier-SLM & 74.4 & 2627 & 0 & 2627 & 0 \\
L1-SLM* & 72.4 & 3020 & 0 & 3020 & 0 \\
\cellcolor{gray!15}Baseline LLM & \cellcolor{gray!15}81.2 & \cellcolor{gray!15}0 & \cellcolor{gray!15}11499 & \cellcolor{gray!15}11499 & \cellcolor{gray!15}100 \\
\midrule
\multicolumn{6}{l}{\textbf{\textit{Collaborative}}} \\
SLM-Verb & 77.2 {\scriptsize (-4.0)} & 3113 & 5409 & 8522 {\scriptsize (-25.9\%)} & 42.1 \\
RLVR-SLM-Verb & 70.8 {\scriptsize (-10.4)} & 2125 & 212 & 2337 {\scriptsize (-79.7\%)} & 1.5 \\
RLVR-SLM-AvgProb & 70.5 {\scriptsize (-10.7)} & 2125 & 0 & 2125 {\scriptsize (-81.5\%)} & 0 \\
RLVR-SLM-Probe & 77.5 {\scriptsize (-3.7)} & 2125 & 5126 & 7252 {\scriptsize (-36.9\%)} & 32.4 \\
Router+RLVR-SLM & 74.4 {\scriptsize (-6.8)} & 1459 & 4031 & 5603 {\scriptsize (-51.3\%)} & 25.8 \\
Brier-SLM-Verb & 81.0 {\scriptsize (-0.2)} & 2627 & 8705 & 11332 {\scriptsize (-1.5\%)} & 63.6 \\
\model* & 80.0 {\scriptsize (-1.2)} & 3020 & 6591 & 9610 {\scriptsize (-16.4\%)} & 43.1 \\
\bottomrule
\end{tabular}
}
\caption{Performance comparison across different systems on Math500. Pass@1, SLM Cost per question, LLM Cost per question, Total Cost per question, and LLM usage percentage (LLM\%) are reported. In all collaborative systems, the Baseline LLM's Pass@1 serves as the confidence threshold. Values in parentheses indicate the difference from the Baseline LLM.  \model (L1-SLM-Verb) utilizes L1-SLM and verbalized confidence. * indicates our method.}
\label{tab:math500}
\end{table*}

\begin{table*}[t]
\centering
\resizebox{0.8\textwidth}{!}{
\begin{tabular}{lccccc}
\toprule
\textbf{System} & \textbf{Pass@1\%} & \textbf{SLM Cost} & \textbf{LLM Cost} & \textbf{Total Cost} & \textbf{LLM\%} \\
\midrule
\multicolumn{6}{l}{\textbf{\textit{Standalone}}} \\
SLM & 81.7 & 1388 & 0 & 1388 & 0 \\
RLVR-SLM & 90.9 & 1348 & 0 & 1348 & 0 \\
Brier-SLM & 90.9 & 1369 & 0 & 1369 & 0 \\
L1-SLM* & 90.3 & 1487 & 0 & 1487 & 0 \\
\cellcolor{gray!15}Baseline LLM & \cellcolor{gray!15}95.6 & \cellcolor{gray!15}0 & \cellcolor{gray!15}6073 & \cellcolor{gray!15}6073 & \cellcolor{gray!15}100 \\
\midrule
\multicolumn{6}{l}{\textbf{\textit{Collaborative}}} \\
SLM-Verb & 92.1 {\scriptsize (-3.6)} & 1388 & 1280 & 2667 {\scriptsize (-56.1\%)} & 19.4 \\
RLVR-SLM-Verb & 92.1 {\scriptsize (-3.5)} & 1348 & 1623 & 2972 {\scriptsize (-51.1\%)} & 25.1 \\
RLVR-SLM-AvgProb & 92.2 {\scriptsize (-3.4)} & 1348 & 525 & 1874 {\scriptsize (-69.1\%)} & 8.4 \\
RLVR-SLM-Probe & 92.8 {\scriptsize (-2.9)} & 1348 & 646 & 1994 {\scriptsize (-67.2\%)} & 9.0 \\
Router+RLVR-SLM & 91.2 {\scriptsize (-4.4)} & 1324 & 104 & 1532 {\scriptsize (-74.8\%)} & 1.5 \\
Brier-SLM-Verb & 95.4 {\scriptsize (-0.3)} & 1369 & 3392 & 4761 {\scriptsize (-21.6\%)} & 51.1 \\
\model* & 94.2 {\scriptsize (-1.4)} & 1487 & 1204 & 2691 {\scriptsize (-55.7\%)} & 17.1 \\
\bottomrule
\end{tabular}
}
\caption{Performance comparison across different systems on GSM8K. Pass@1, SLM Cost per question, LLM Cost per question, Total Cost per question, and LLM usage percentage (LLM\%) are reported. In all collaborative systems, the Baseline LLM's Pass@1 serves as the confidence threshold. Values in parentheses indicate the difference from the Baseline LLM.  \model (L1-SLM-Verb) utilizes L1-SLM and verbalized confidence. * indicates our method.}
\label{tab:main_results_gsm8k}
\end{table*}

\begin{table*}[t]
\centering
\resizebox{0.8\textwidth}{!}{
\begin{tabular}{lccccc}
\toprule
\textbf{System} & \textbf{Pass@1\%} & \textbf{SLM Cost} & \textbf{LLM Cost} & \textbf{Total Cost} & \textbf{LLM\%} \\
\midrule
\multicolumn{6}{l}{\textbf{\textit{Standalone}}} \\
SLM & 23.4 & 4671 & 0 & 4671 & 0 \\
RLVR-SLM & 35.2 & 2916 & 0 & 2916 & 0 \\
Brier-SLM & 36.0 & 3960 & 0 & 3960 & 0 \\
L1-SLM* & 36.5 & 4424 & 0 & 4424 & 0 \\
\cellcolor{gray!15}Baseline LLM & \cellcolor{gray!15}47.0 & \cellcolor{gray!15}0 & \cellcolor{gray!15}17167 & \cellcolor{gray!15}17167 & \cellcolor{gray!15}100 \\
\midrule
\multicolumn{6}{l}{\textbf{\textit{Collaborative}}} \\
SLM-Verb & 42.5 {\scriptsize (-4.5)} & 4671 & 10778 & 15449 {\scriptsize (-10.0\%)} & 59.8 \\
RLVR-SLM-Verb & 35.7 {\scriptsize (-11.3)} & 2916 & 361 & 3277 {\scriptsize (-80.9\%)} & 1.4 \\
RLVR-SLM-AvgProb & 35.2 {\scriptsize (-11.8)} & 2916 & 0 & 2916 {\scriptsize (-83.0\%)} & 0 \\
RLVR-SLM-Probe & 41.6 {\scriptsize (-5.5)} & 2916 & 9519 & 12434 {\scriptsize (-27.6\%)} & 51.4 \\
Router+RLVR-SLM & 41.6 {\scriptsize (-5.4)} & 1293 & 9800 & 11236 {\scriptsize (-34.6\%)} & 54.2 \\
Brier-SLM-Verb & 44.9 {\scriptsize (-2.1)} & 3960 & 12572 & 16532 {\scriptsize (-3.7\%)} & 65.2 \\
\model* & 44.3 {\scriptsize (-2.7)} & 4424 & 12677 & 17101 {\scriptsize (-0.4\%)} & 65.6 \\
\bottomrule
\end{tabular}
}
\caption{Performance comparison across different systems on OlympiadBench. Pass@1, SLM Cost per question, LLM Cost per question, Total Cost per question, and LLM usage percentage (LLM\%) are reported. In all collaborative systems, the Baseline LLM's Pass@1 serves as the confidence threshold. Values in parentheses indicate the difference from the Baseline LLM.  \model (L1-SLM-Verb) utilizes L1-SLM and verbalized confidence. * indicates our method.}
\label{tab:main_results_olympiadbench}
\end{table*}

\begin{table*}[t]
\centering
\resizebox{0.8\textwidth}{!}{
\begin{tabular}{lccccc}
\toprule
\textbf{System} & \textbf{Pass@1\%} & \textbf{SLM Cost} & \textbf{LLM Cost} & \textbf{Total Cost} & \textbf{LLM\%} \\
\midrule
\multicolumn{6}{l}{\textbf{\textit{Standalone}}} \\
SLM & 21.6 & 3927 & 0 & 3927 & 0 \\
RLVR-SLM & 29.5 & 2451 & 0 & 2451 & 0 \\
Brier-SLM & 32.8 & 3282 & 0 & 3282 & 0 \\
L1-SLM* & 34.2 & 3775 & 0 & 3775 & 0 \\
\cellcolor{gray!15}Baseline LLM & \cellcolor{gray!15}43.4 & \cellcolor{gray!15}0 & \cellcolor{gray!15}13420 & \cellcolor{gray!15}13420 & \cellcolor{gray!15}100 \\
\midrule
\multicolumn{6}{l}{\textbf{\textit{Collaborative}}} \\
SLM-Verb & 22.9 {\scriptsize (-20.4)} & 3927 & 630 & 4557 {\scriptsize (-66.0\%)} & 4.1 \\
RLVR-SLM-Verb & 29.7 {\scriptsize (-13.7)} & 2451 & 108 & 2559 {\scriptsize (-80.9\%)} & 0.6 \\
RLVR-SLM-AvgProb & 29.5 {\scriptsize (-13.9)} & 2451 & 0 & 2451 {\scriptsize (-81.7\%)} & 0 \\
RLVR-SLM-Probe & 39.7 {\scriptsize (-3.7)} & 2451 & 9302 & 11753 {\scriptsize (-12.4\%)} & 65.5 \\
Router+RLVR-SLM & 35.1 {\scriptsize (-8.3)} & 1407 & 5810 & 7463 {\scriptsize (-44.4\%)} & 42.6 \\
Brier-SLM-Verb & 39.0 {\scriptsize (-4.3)} & 3282 & 7808 & 11090 {\scriptsize (-17.4\%)} & 54.0 \\
\model* & 38.2 {\scriptsize (-5.1)} & 3775 & 6734 & 10509 {\scriptsize (-21.7\%)} & 45.8 \\
\bottomrule
\end{tabular}
}
\caption{Performance comparison across different systems on GPQA. Pass@1, SLM Cost per question, LLM Cost per question, Total Cost per question, and LLM usage percentage (LLM\%) are reported. In all collaborative systems, the Baseline LLM's Pass@1 serves as the confidence threshold. Values in parentheses indicate the difference from the Baseline LLM.  \model (L1-SLM-Verb) utilizes L1-SLM and verbalized confidence. * indicates our method.}
\label{tab:main_results_gpqa}
\end{table*}

\begin{table*}[t]
\centering
\resizebox{0.8\textwidth}{!}{
\begin{tabular}{lccccc}
\toprule
\textbf{System} & \textbf{Pass@1\%} & \textbf{SLM Cost} & \textbf{LLM Cost} & \textbf{Total Cost} & \textbf{LLM\%} \\
\midrule
\multicolumn{6}{l}{\textbf{\textit{Standalone}}} \\
SLM & 53.9 & 1298 & 0 & 1298 & 0 \\
RLVR-SLM & 74.5 & 1067 & 0 & 1067 & 0 \\
Brier-SLM & 74.8 & 1130 & 0 & 1130 & 0 \\
L1-SLM* & 75.9 & 1215 & 0 & 1215 & 0 \\
\cellcolor{gray!15}Baseline LLM & \cellcolor{gray!15}81.7 & \cellcolor{gray!15}0 & \cellcolor{gray!15}5229 & \cellcolor{gray!15}5229 & \cellcolor{gray!15}100 \\
\midrule
\multicolumn{6}{l}{\textbf{\textit{Collaborative}}} \\
SLM-Verb & 58.3 {\scriptsize (-23.4)} & 1298 & 597 & 1895 {\scriptsize (-63.8\%)} & 11.4 \\
RLVR-SLM-Verb & 74.8 {\scriptsize (-6.9)} & 1067 & 58 & 1125 {\scriptsize (-78.5\%)} & 1.3 \\
RLVR-SLM-AvgProb & 74.6 {\scriptsize (-7.1)} & 1067 & 68 & 1134 {\scriptsize (-78.3\%)} & 1.2 \\
RLVR-SLM-Probe & 77.4 {\scriptsize (-4.3)} & 1067 & 1079 & 2145 {\scriptsize (-59.0\%)} & 20.6 \\
Router+RLVR-SLM & 76.5 {\scriptsize (-5.2)} & 737 & 1614 & 2431 {\scriptsize (-53.5\%)} & 30.9 \\
Brier-SLM-Verb & 81.7 {\scriptsize (+0.0)} & 1130 & 4626 & 5756 {\scriptsize (+10.1\%)} & 88.1 \\
\model* & 81.3 {\scriptsize (-0.4)} & 1215 & 3373 & 4589 {\scriptsize (-12.3\%)} & 63.8 \\
\bottomrule
\end{tabular}
}
\caption{Performance comparison across different systems on CommonsenseQA. Pass@1, SLM Cost per question, LLM Cost per question, Total Cost per question, and LLM usage percentage (LLM\%) are reported. In all collaborative systems, the Baseline LLM's Pass@1 serves as the confidence threshold. Values in parentheses indicate the difference from the Baseline LLM.  \model (L1-SLM-Verb) utilizes L1-SLM and verbalized confidence. * indicates our method.}
\label{tab:main_results_commonsenseqa}
\end{table*}

Tables \ref{tab:main_results_deepmath500}–\ref{tab:main_results_commonsenseqa} compare the performance of different systems on DeepMath500, Math500, GSM8K, OlympiadBench, GPQA, and CommonsenseQA, respectively. The confidence threshold is set to match the Baseline LLM's pass@1 for each dataset. The results align with our main paper findings: except for GPQA, \model achieves significant cost reductions with only marginal accuracy drops compared to the baseline LLM. Compared to other collaborative systems, \model achieves a superior accuracy-cost balance. On GPQA, the pass@1 drop is slightly larger (-5.1\%) due to less well-calibrated confidence scores.

\section{A Different SLM+LLM Paring}
\label{sec:slm1.5b_llm32b}

\begin{table*}[htbp]
\centering
\resizebox{\textwidth}{!}{
\begin{tabular}{lccccccccc}
\toprule[1.5pt]
\multirow{2}{*}{\textbf{System}} & \multicolumn{3}{c}{\textbf{DeepMath500}} & \multicolumn{3}{c}{\textbf{OOD Math}} & \multicolumn{3}{c}{\textbf{OOD Non-Math}} \\
\cmidrule(lr){2-4} \cmidrule(lr){5-7} \cmidrule(lr){8-10}
& \textbf{Pass@1\%} ↑ & \textbf{Avg Cost} ↓ & \textbf{LLM\%} & \textbf{Pass@1\%} ↑ & \textbf{Avg Cost} ↓ & \textbf{LLM\%} & \textbf{Pass@1\%} ↑ & \textbf{Avg Cost} ↓ & \textbf{LLM\%} \\
\midrule[1pt]
\multicolumn{10}{l}{\textbf{\textit{Standalone}}} \\
SLM & 25.8 & 3422 & 0.0 & 41.5 & 2435 & 0.0 & 42.8 & 1558 & 0.0 \\
L1-SLM & 26.5 & 1311 & 0.0 & 35.3 & 1273 & 0.0 & 35.7 & 449 & 0.0 \\
\cellcolor{gray!15}Baseline LLM & \cellcolor{gray!15}69.0 & \cellcolor{gray!15}69450 & \cellcolor{gray!15}100.0 & \cellcolor{gray!15}79.6 & \cellcolor{gray!15}47430 & \cellcolor{gray!15}100.0 & \cellcolor{gray!15}71.4 & \cellcolor{gray!15}34663 & \cellcolor{gray!15}100.0 \\
\midrule
\multicolumn{10}{l}{\textbf{\textit{Collaborative}}} \\
SLM-Verb & 68.7\,{\scriptsize(--0.3)} & 72495\,{\scriptsize(+4.4\%)} & 99.3 & 79.1\,{\scriptsize(--0.5)} & 49692\,{\scriptsize(+4.8\%)} & 99.0 & 71.3\,{\scriptsize(--0.1)} & 35492\,{\scriptsize(+2.4\%)} & 98.4 \\
\model* & 55.7\,{\scriptsize(--13.2)} & 46490\,{\scriptsize(--33.1\%)} & 63.1 & 70.7\,{\scriptsize(--8.9)} & 28996\,{\scriptsize(--38.9\%)} & 75.2 & 67.6\,{\scriptsize(--3.8)} & 19977\,{\scriptsize(--42.4\%)} & 76.2 \\
\bottomrule[1.5pt]
\end{tabular}
}
\caption{Performance comparison of 1.5B SLM + 32B LLM pairing across different systems and datasets. Pass@1, average cost per question (Avg Cost), and LLM usage percentage (LLM\%) are reported. OOD Math includes Math500, GSM8K, and OlympiadBench. OOD Non-Math includes GPQA and CommonsenseQA. In all collaborative systems, the Baseline LLM's Pass@1 serves as the confidence threshold. Values in parentheses indicate the difference from the Baseline LLM. * indicates our method.}
\label{tab:1.5b_32b_results}
\end{table*}

In this section, we evaluate a 1.5B + 32B model pairing. Specifically, we utilize Qwen2.5-1.5B-Instruct as the base SLM and Qwen2.5-32B-Instruct as the base LLM, training both with the RLCC-L1 algorithm. Results are presented in Table~\ref{tab:1.5b_32b_results}.

We select the checkpoint at step 30 rather than step 150, as we observed that the SLM gradually becomes underconfident during training, eventually returning 0 as the confidence score in most cases. At step 30, the model achieves better confidence calibration while maintaining comparable reasoning ability.

As shown, \model~remains effective for OOD Non-Math, achieving substantial cost reduction (--42.4\%) with only a modest accuracy drop (--3.8\%). However, compared to the 7B+32B results (Table~\ref{tab:main_results}), the 1.5B+32B system suffers more substantial performance degradation on DeepMath500 and OOD Math despite achieving greater cost reductions. Analysis reveals that the 1.5B SLM's confidence is more poorly calibrated than that of the 7B model---specifically, it tends to exhibit overconfidence on problems beyond its capability. This suggests that beyond weaker reasoning ability, smaller models may also demonstrate diminished self-awareness. We plan to investigate the scaling laws of model self-awareness in future work.

\section{Detailed Confidence Calibration Results}
\label{sec:detailed_calibration}

\begin{figure*}[htbp]
\centering
\begin{subfigure}[t]{0.48\textwidth}
\centering
\includegraphics[width=\linewidth]{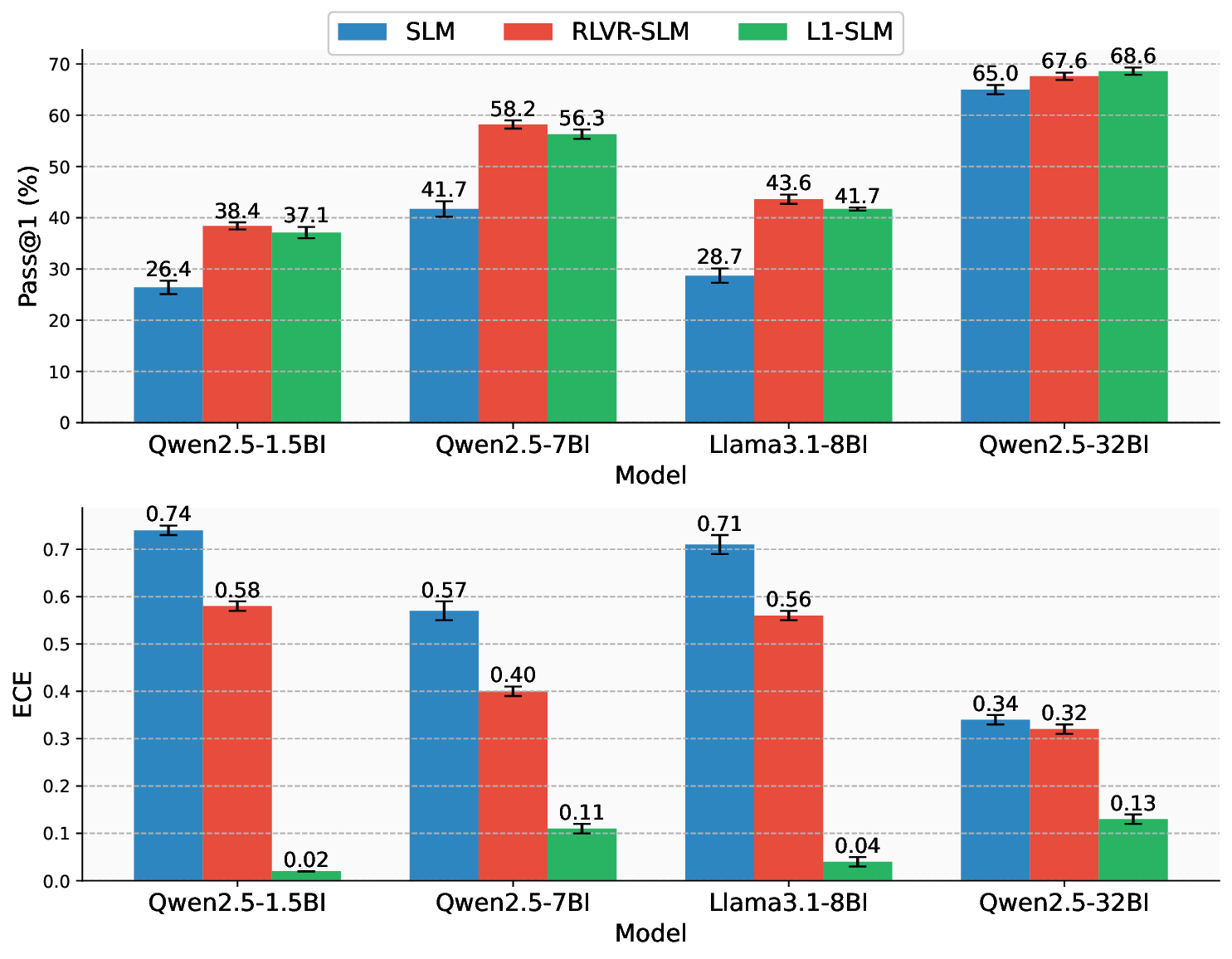}
\caption{DeepMath500}
\label{fig:model_comparison_bars_deepmath500}
\end{subfigure}
\hfill
\begin{subfigure}[t]{0.48\textwidth}
\centering
\includegraphics[width=\linewidth]{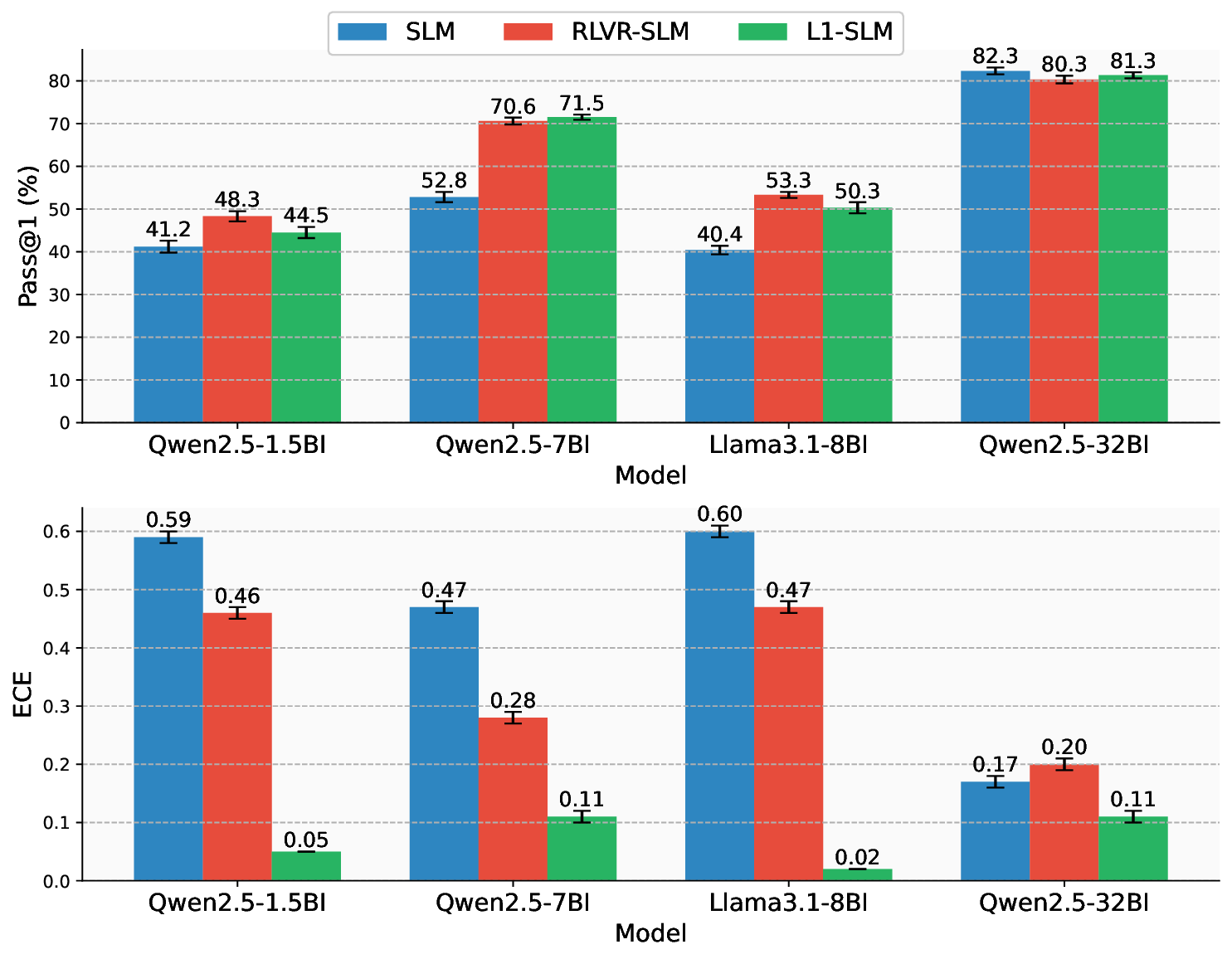}
\caption{Math500}
\label{fig:model_comparison_bars_math500}
\end{subfigure}
\begin{subfigure}[t]{0.48\textwidth}
\centering
\includegraphics[width=\linewidth]{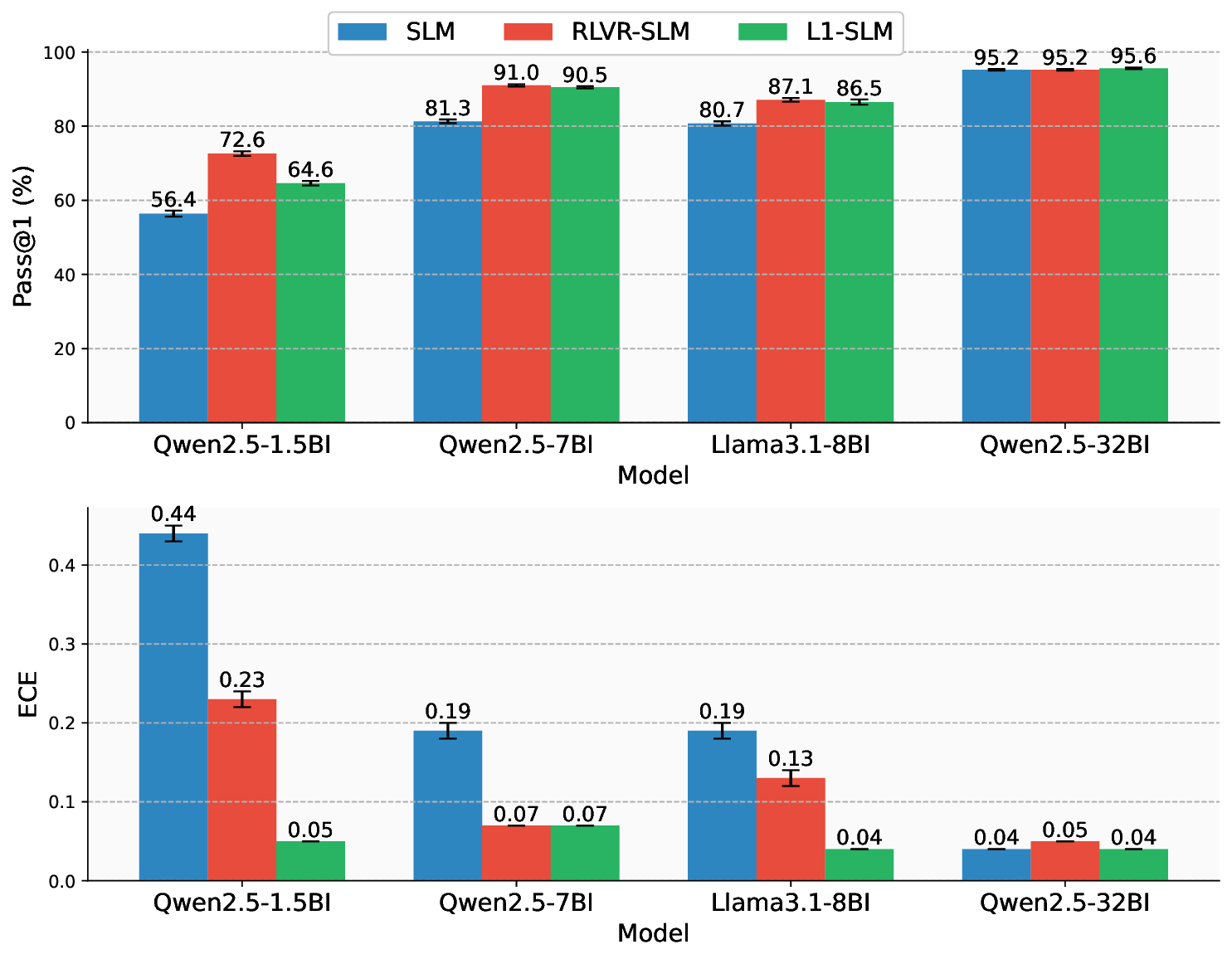}
\caption{GSM8K}
\label{fig:model_comparison_bars_gsm8k}
\end{subfigure}
\begin{subfigure}[t]{0.48\textwidth}
\centering
\includegraphics[width=\linewidth]{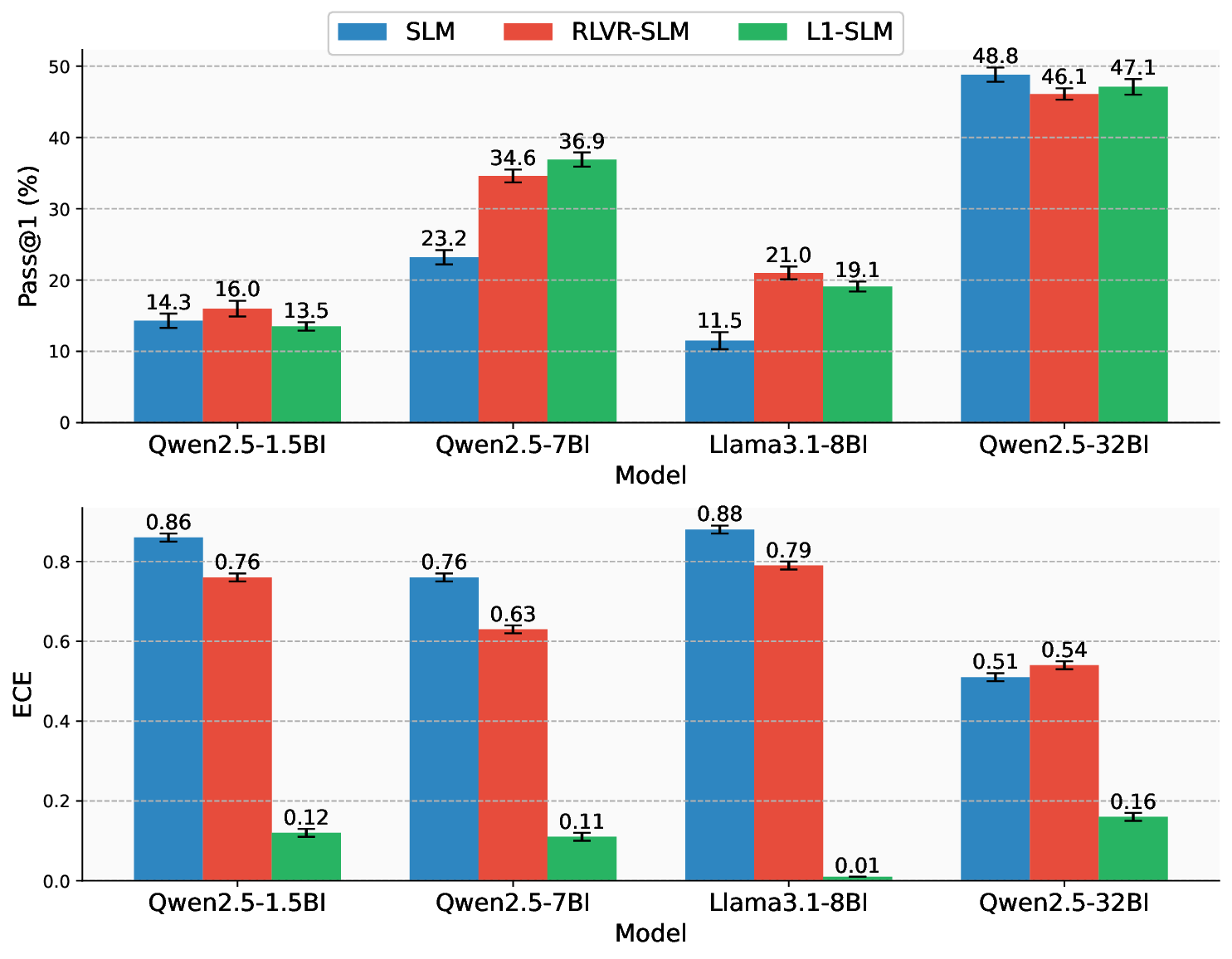}
\caption{OlympiadBench}
\label{fig:model_comparison_bars_olympiadbench}
\end{subfigure}
\begin{subfigure}[t]{0.48\textwidth}
\centering
\includegraphics[width=\linewidth]{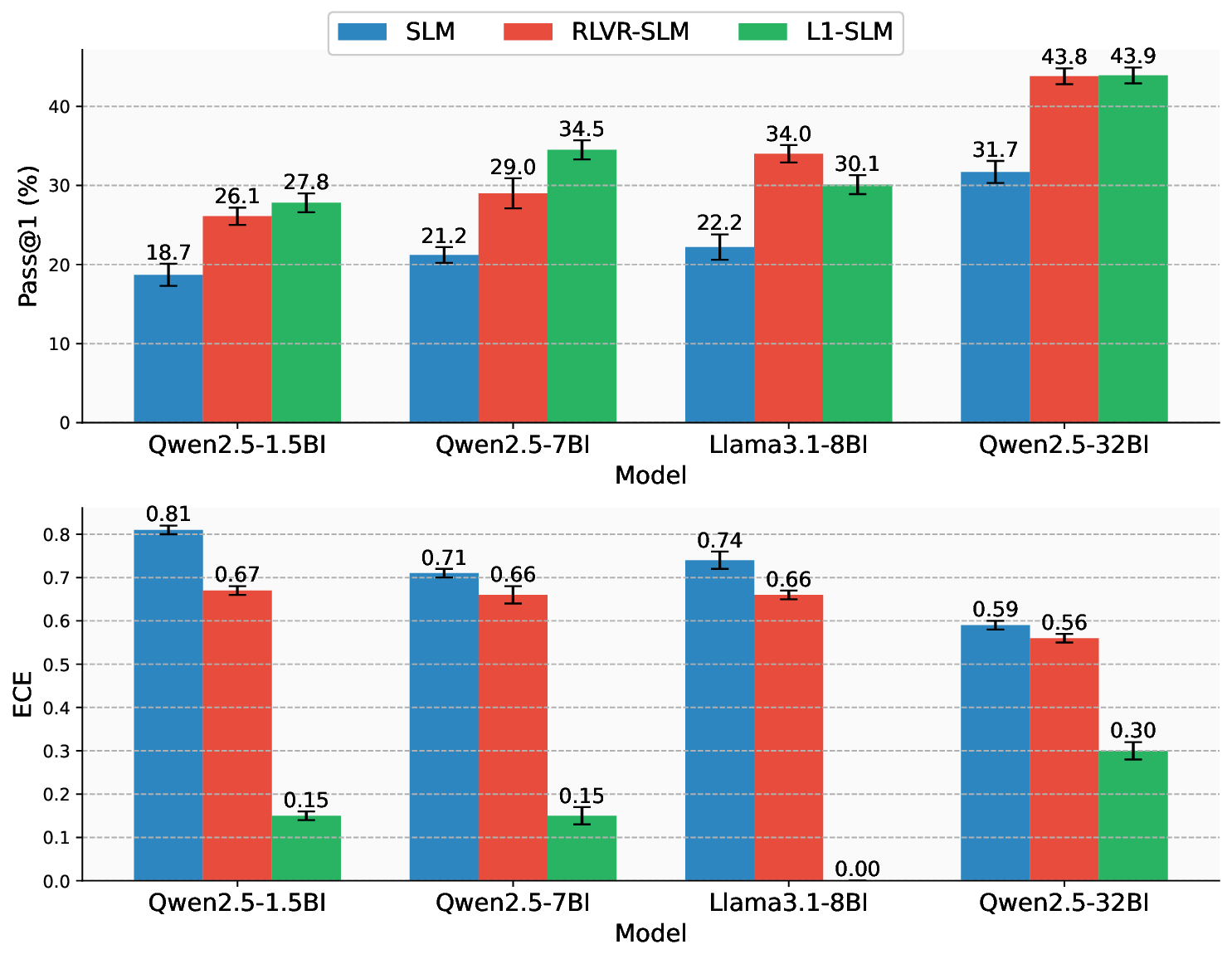}
\caption{GPQA}
\label{fig:model_comparison_bars_gpqa}
\end{subfigure}
\begin{subfigure}[t]{0.48\textwidth}
\centering
\includegraphics[width=\linewidth]{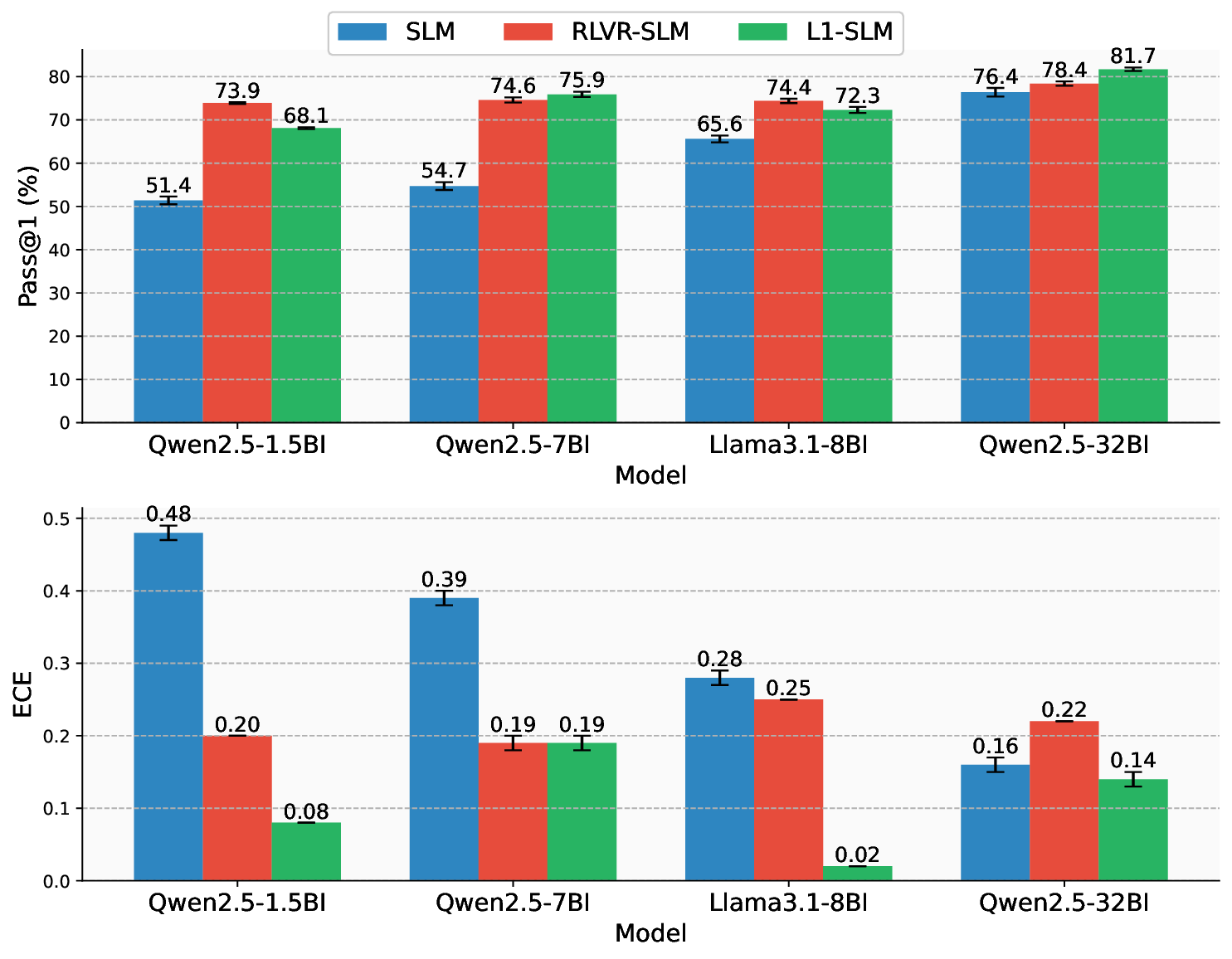}
\caption{CommonsenseQA}
\label{fig:model_comparison_bars_commonsenseqa}
\end{subfigure}
\caption{Comparison of Pass@1 and ECE across several models trained with different reward configurations on DeepMath500, Math500, GSM8K, OlympiadBench, GPQA, OlympiadBench, respectively.}
\label{fig:model_comparisons_datasets}
\end{figure*}

Figure~\ref{fig:model_comparisons_datasets} shows the Pass@1 accuracy and Expected Calibration Error (ECE) of different base models on each evaluation dataset. These results indicate that RL training effectively improves the model's reasoning capability across all in-domain and out-of-domain evaluation datasets, with the exception of the Qwen2.5-32BI model, whose reasoning ability is already sufficiently strong that further improvement is limited.

However, RL training without confidence reward fails to calibrate the model's confidence predictions with its true accuracy performance, and in some cases even degrades calibration, as observed in the Math500, OlympiadBench, and CommonsenseQA with Qwen2.5-32BI. In contrast, incorporating a confidence reward into RL training enables the model to effectively align its confidence with true accuracy, as evidenced by significant ECE reductions across all base models and datasets.

\section{Results of Different Confidence Reward Functions on Each Dataset}
\label{sec:different_confidence_reward_dataset}

\begin{table*}[t]
\centering
\resizebox{0.98\textwidth}{!}{%
\begin{tabular}{cccccccccc}
\toprule
\multirow{2}{*}{\textbf{Method}} & \multicolumn{3}{c}{\textbf{DeepMath500}} & \multicolumn{3}{c}{\textbf{Math500}} & \multicolumn{3}{c}{\textbf{GSM8K}} \\
\cmidrule(lr){2-4} \cmidrule(lr){5-7} \cmidrule(lr){8-10}
& \textbf{Pass@1\%} & \textbf{ECE} & \textbf{AUROC} & \textbf{Pass@1\%} & \textbf{ECE} & \textbf{AUROC} & \textbf{Pass@1\%} & \textbf{ECE} & \textbf{AUROC} \\
\midrule
SLM         & 41.7±1.5 & 0.57±0.02 & 0.47±0.02 & 52.8±1.2 & 0.47±0.01 & 0.48±0.02 & 81.3±0.5 & 0.19±0.01 & 0.52±0.00 \\
RLVR        & 58.2±0.8 & 0.40±0.01 & 0.49±0.01 & 70.6±0.8 & 0.28±0.01 & 0.50±0.01 & 91.0±0.3 & 0.07±0.00 & 0.49±0.01 \\
L1          & 56.3±0.9 & 0.11±0.01 & 0.77±0.01 & 71.5±0.6 & 0.11±0.01 & 0.81±0.01 & 90.5±0.3 & 0.07±0.00 & 0.78±0.01 \\
Sample\_L1  & 58.4±0.8 & 0.17±0.01 & 0.75±0.01 & 73.6±0.9 & 0.07±0.01 & 0.81±0.01 & 91.1±0.3 & 0.05±0.00 & 0.70±0.01 \\
L2          & 54.8±1.1 & 0.22±0.02 & 0.63±0.01 & 72.6±0.9 & 0.39±0.01 & 0.67±0.01 & 89.7±0.2 & 0.53±0.00 & 0.67±0.01 \\
Brier       & 55.9±1.1 & 0.12±0.01 & 0.73±0.01 & 73.8±1.1 & 0.15±0.02 & 0.81±0.01 & 90.8±0.2 & 0.20±0.00 & 0.80±0.01 \\
KL         & 56.2±0.8 & 0.12±0.02 & 0.73±0.02 & 71.5±1.2 & 0.16±0.01 & 0.76±0.02 & 90.3±0.4 & 0.23±0.00 & 0.77±0.01 \\
Sample\_KL & 54.9±1.2 & 0.13±0.01 & 0.75±0.01 & 74.0±0.7 & 0.11±0.01 & 0.86±0.01 & 90.3±0.5 & 0.08±0.00 & 0.81±0.01 \\
\bottomrule
\end{tabular}
}
\caption{Comparison of different confidence reward functions on accuracy and calibration metrics for DeepMath500, Math500, and GSM8K. Results are reported as mean ± standard deviation.}
\label{tab:conf_reward_deepmath500_math500_gsm8k}
\end{table*}

\begin{table*}[t]
\centering
\resizebox{0.98\textwidth}{!}{%
\begin{tabular}{cccccccccc}
\toprule
\multirow{2}{*}{\textbf{Method}} & \multicolumn{3}{c}{\textbf{OlympiadBench}} & \multicolumn{3}{c}{\textbf{GPQA}} & \multicolumn{3}{c}{\textbf{CommonsenseQA}} \\
\cmidrule(lr){2-4} \cmidrule(lr){5-7} \cmidrule(lr){8-10}
& \textbf{Pass@1\%} & \textbf{ECE} & \textbf{AUROC} & \textbf{Pass@1\%} & \textbf{ECE} & \textbf{AUROC} & \textbf{Pass@1\%} & \textbf{ECE} & \textbf{AUROC} \\
\midrule
SLM         & 23.2±1.0 & 0.76±0.01 & 0.48±0.01 & 21.2±1.0 & 0.71±0.01 & 0.49±0.03 & 54.7±0.9 & 0.39±0.01 & 0.52±0.01 \\
RLVR        & 34.6±0.9 & 0.63±0.01 & 0.49±0.01 & 29.0±1.9 & 0.66±0.02 & 0.50±0.03 & 74.6±0.6 & 0.19±0.01 & 0.52±0.01 \\
L1          & 36.9±1.0 & 0.11±0.01 & 0.77±0.01 & 34.5±1.2 & 0.15±0.02 & 0.54±0.02 & 75.9±0.6 & 0.19±0.01 & 0.63±0.01 \\
Sample\_L1  & 37.5±1.4 & 0.22±0.01 & 0.74±0.01 & 32.3±0.8 & 0.40±0.01 & 0.56±0.02 & 75.8±0.6 & 0.18±0.01 & 0.60±0.00 \\
L2          & 36.5±0.5 & 0.10±0.00 & 0.60±0.01 & 32.2±1.6 & 0.04±0.01 & 0.54±0.02 & 76.9±0.9 & 0.28±0.01 & 0.60±0.01 \\
Brier       & 35.6±0.8 & 0.10±0.01 & 0.77±0.01 & 32.9±1.2 & 0.11±0.01 & 0.54±0.02 & 74.9±0.6 & 0.22±0.01 & 0.62±0.01 \\
KL         & 35.4±0.7 & 0.09±0.01 & 0.68±0.01 & 30.9±1.2 & 0.14±0.01 & 0.55±0.01 & 73.3±0.7 & 0.12±0.01 & 0.61±0.01 \\
Sample\_KL & 36.3±1.2 & 0.09±0.01 & 0.82±0.01 & 32.8±0.9 & 0.18±0.01 & 0.55±0.02 & 73.8±0.5 & 0.12±0.01 & 0.63±0.01 \\
\bottomrule
\end{tabular}
}
\caption{Comparison of different confidence reward functions on accuracy and calibration metrics for OlympiadBench, GPQA, and CommonsenseQA. Results are reported as mean ± standard deviation.}
\label{tab:conf_reward_others}
\end{table*}

Tables \ref{tab:conf_reward_deepmath500_math500_gsm8k} and \ref{tab:conf_reward_others} present detailed accuracy and calibration results for each dataset. As observed, except for L2, all other confidence reward functions achieve comparable pass@1, ECE, and AUROC across the six datasets. This indicates that L1, Sample\_L1, Brier, KL, and Sample\_KL are all effective in the RL training algorithm. Among these functions, we recommend L1 as it demonstrates the strongest balance between accuracy and calibration.

\section{Rollout Size Results on Each Dataset}
\label{sec:rollout_dataset}

\begin{table*}[t]
\centering
\resizebox{0.98\textwidth}{!}{%
\begin{tabular}{cccccccccc}
\toprule
\multirow{2}{*}{\textbf{Rollout Size}} & \multicolumn{3}{c}{\textbf{DeepMath500}} & \multicolumn{3}{c}{\textbf{Math500}} & \multicolumn{3}{c}{\textbf{GSM8K}} \\
\cmidrule(lr){2-4} \cmidrule(lr){5-7} \cmidrule(lr){8-10}
& \textbf{Pass@1\%} & \textbf{ECE} & \textbf{AUROC} & \textbf{Pass@1\%} & \textbf{ECE} & \textbf{AUROC} & \textbf{Pass@1\%} & \textbf{ECE} & \textbf{AUROC} \\
\midrule
4  & 55.8±1.0 & 0.12±0.01 & 0.67±0.01 & 71.2±1.0 & 0.13±0.01 & 0.74±0.02 & 91.7±0.4 & 0.07±0.01 & 0.73±0.02 \\
8  & 56.3±0.9 & 0.11±0.01 & 0.77±0.01 & 71.5±0.6 & 0.11±0.01 & 0.81±0.01 & 90.5±0.3 & 0.07±0.00 & 0.78±0.01 \\
16 & 54.7±1.1 & 0.09±0.01 & 0.69±0.01 & 68.9±1.3 & 0.08±0.01 & 0.74±0.01 & 91.3±0.3 & 0.07±0.00 & 0.74±0.00 \\
32 & 57.2±0.9 & 0.19±0.01 & 0.73±0.01 & 72.1±0.5 & 0.11±0.01 & 0.78±0.02 & 91.7±0.2 & 0.06±0.00 & 0.62±0.01 \\
\bottomrule
\end{tabular}
}
\caption{Comparison of different rollout sizes with L1 confidence reward on accuracy and calibration metrics for DeepMath500, Math500, and GSM8K. Results are reported as mean ± standard deviation.}
\label{tab:rollout_analysis_deepmath500_math500_gsm8k}
\end{table*}

\begin{table*}[t]
\centering
\resizebox{0.98\textwidth}{!}{%
\begin{tabular}{cccccccccc}
\toprule
\multirow{2}{*}{\textbf{Rollout Size}} & \multicolumn{3}{c}{\textbf{OlympiadBench}} & \multicolumn{3}{c}{\textbf{GPQA}} & \multicolumn{3}{c}{\textbf{CommonsenseQA}} \\
\cmidrule(lr){2-4} \cmidrule(lr){5-7} \cmidrule(lr){8-10}
& \textbf{Pass@1\%} & \textbf{ECE} & \textbf{AUROC} & \textbf{Pass@1\%} & \textbf{ECE} & \textbf{AUROC} & \textbf{Pass@1\%} & \textbf{ECE} & \textbf{AUROC} \\
\midrule
4  & 35.2±0.9 & 0.07±0.01 & 0.63±0.02 & 29.0±1.1 & 0.20±0.01 & 0.56±0.02 & 75.4±0.9 & 0.11±0.01 & 0.69±0.01 \\
8  & 36.9±1.0 & 0.11±0.01 & 0.77±0.01 & 34.5±1.2 & 0.15±0.02 & 0.54±0.02 & 75.9±0.6 & 0.19±0.01 & 0.63±0.01 \\
16 & 33.3±1.0 & 0.08±0.01 & 0.69±0.01 & 31.7±1.2 & 0.24±0.01 & 0.55±0.02 & 75.6±0.5 & 0.09±0.01 & 0.66±0.01 \\
32 & 34.2±0.7 & 0.20±0.01 & 0.80±0.01 & 34.0±1.4 & 0.42±0.02 & 0.53±0.02 & 77.5±0.3 & 0.21±0.00 & 0.55±0.01 \\
\bottomrule
\end{tabular}
}
\caption{Comparison of different rollout sizes with L1 confidence reward on accuracy and calibration metrics for OlympiadBench, GPQA, and CommonsenseQA. Results are reported as mean ± standard deviation.}
\label{tab:rollout_analysis_other}
\end{table*}

Tables \ref{tab:rollout_analysis_deepmath500_math500_gsm8k} and \ref{tab:rollout_analysis_other} present detailed results across different rollout sizes for each dataset. The metric differences are marginal when increasing rollout size from 4 to 32, indicating that our RL training method is robust to rollout size selection. From a training efficiency perspective, we recommend using a rollout size of 4 or 8.

\end{document}